\documentclass[lettersize,journal]{IEEEtran}
\usepackage{amsmath,amsfonts}
\usepackage{graphicx}
\usepackage{amsmath}
\usepackage{amssymb}
\usepackage{booktabs}
\usepackage{bm}
\usepackage{multirow}
\usepackage{times}
\usepackage{makecell}
\usepackage{enumitem}
\usepackage{subfigure}
\usepackage{amsthm}
\usepackage{mathrsfs}
\usepackage{color}
\usepackage[pagebackref=true,breaklinks=true,colorlinks,bookmarks=false,citecolor=blue,linkcolor=blue]{hyperref}
\usepackage{hyperref}

\begin{document}

\title{Recurrent Generic Contour-based Instance Segmentation with Progressive Learning}

\author{Hao Feng$^{\dagger}$,
 Keyi Zhou$^{\dagger}$,
 Wengang Zhou*,~\IEEEmembership{Senior Member,~IEEE}, \par
 Yufei Yin,
 Jiajun Deng,
 Qi Sun,
 and~Houqiang Li*,~\IEEEmembership{Fellow,~IEEE}
 
\IEEEcompsocitemizethanks{
\IEEEcompsocthanksitem Hao Feng, Keyi Zhou, Wengang Zhou, Yufei Yin, Qi Sun, and Houqiang Li are with the CAS Key Laboratory of Technology in Geo-spatial Information Processing and Application System, Department of Electronic Engineering and Information Science, University of Science and Technology of China, Hefei, 230027, China.
Hao Feng is also with Zhangjiang Laboratory, Shanghai, China.
E-mail: \{haof, kyzhou2000, yinyufei, sq008\}@mail.ustc.edu.cn; \{zhwg, lihq\}@ustc.edu.cn
\IEEEcompsocthanksitem Jiajun Deng is with The University of Adelaide, Australian Institute for Machine Learning. E-mail: jiajun.deng@adelaide.edu.au
\IEEEcompsocthanksitem $^{\dagger}$The first two authors contribute equally to this work.
\IEEEcompsocthanksitem *Corresponding authors: Wengang Zhou and Houqiang Li.
}}


\maketitle

\begin{abstract}
Contour-based instance segmentation has been actively studied,
thanks to its flexibility and elegance in processing visual objects within complex backgrounds.
In this work, we propose a novel deep network architecture, \emph{i.e.}, PolySnake, 
for generic contour-based instance segmentation. 
Motivated by the classic Snake algorithm, the proposed PolySnake achieves superior and robust segmentation performance with an iterative and progressive contour refinement strategy.
Technically,
PolySnake introduces a recurrent update operator to estimate the object contour iteratively.
It maintains a single estimate of the contour that
is progressively deformed toward the object boundary.
At each iteration,
PolySnake builds a semantic-rich representation for the current contour and feeds it to the recurrent operator for further contour adjustment.
Through the iterative refinements, the contour 
progressively converges to a stable status that tightly encloses the
object instance.
Beyond the scope of general instance segmentation, extensive experiments are conducted to validate the effectiveness and generalizability of our PolySnake in two additional specific task scenarios, including scene text detection and lane detection.
The results demonstrate that the proposed PolySnake outperforms the existing advanced methods
on several multiple prevalent benchmarks across the three tasks.
The codes and pre-trained models are available at \href{https://github.com/fh2019ustc/PolySnake}{{\tt  https://github.com/fh2019ustc/PolySnake}}
\end{abstract}


\section{Introduction}
\IEEEPARstart{I}nstance segmentation is a fundamental computer vision task,  
aiming to recognize each distinct object in an image along with their associated outlines.
It is a challenging task due to background clutter, instance ambiguity, and the complexity of object shapes, \emph{etc}.
The advance of instance segmentation benefits a broad range of visual understanding applications,
such as autonomous driving~\cite{ma2019deep,grigorescu2020survey,jin2022vwp}, augmented reality~\cite{alhaija2017augmented,abu2018augmented}, robotic grasping~\cite{fazeli2019see,kleeberger2020survey}, surface defect detection~\cite{li2021tiny,lin2022emra,huang2022self}, text recognition~\cite{shi2016end,yao2014unified,wang2020r}, and so on.
Over the past few years, instance segmentation has been receiving increased attention and 
significant progress has been made.

\begin{figure}[t]
	\centering
	\includegraphics[width=1\linewidth]{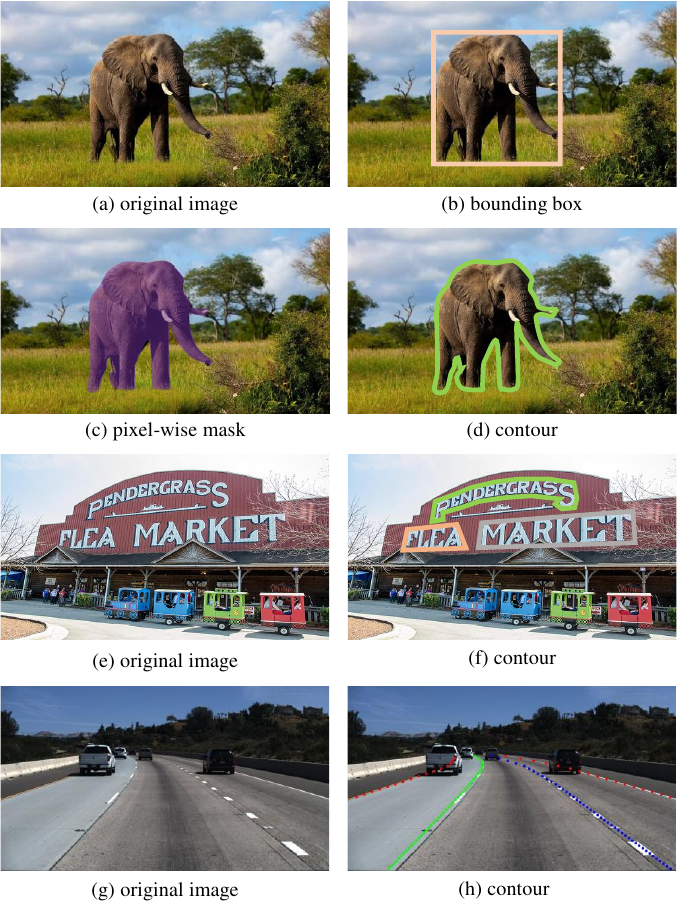}
	\caption{An illustration of different representations to localize an instance within an image: (b) bounding box; (c) pixel-wise mask; and (d), (f), (h) contour. The images showcase three distinct task scenarios: general instance segmentation (a), scene text detection (f), and lane detection (h).
	}
	\label{fig:first}
	\vspace{-0.1in}
\end{figure}

In the literature, most of the state-of-the-art methods~\cite{He_2017_ICCV, chen2019hybrid, Huang_2019_CVPR, liu2018path, shen2021dct, zhang2021refinemask,hu2021a2, tang2021look, minaee2021image} adopt a two-stage pipeline for instance segmentation.
Typically, they first detect the object instance in the form of bounding boxes (see Fig.~\ref{fig:first}(b))
and then estimate a pixel-wise segmentation mask (see Fig.~\ref{fig:first}(c)) within each bounding box.
However, the segmentation performance of such methods is limited due to the inaccurate bounding boxes.
For example, the classic Mask R-CNN~\cite{He_2017_ICCV} performs a binary classification on a 28$\times$28 feature map of a detected instance.
Besides, their dense prediction architecture usually suffers from heavy computational overhead~\cite{shelhamer2017fully}, 
limiting their applications in resource-limited scenarios with a real-time requirement.

To address the above issues, 
recent research efforts have been dedicated to designing alternative representations to the pixel-wise segmentation mask.
One intuitive representative scheme is the object contour, in the form of a sequence of vertexes along the object silhouette (see Fig.~\ref{fig:first}(d)).
Typically,
PolarMask~\cite{Xie_2020_CVPR, xie2021polarmask++} innovatively applies the angle and distance terms in a polar coordinate system to localize the vertices of the contour, which achieves competitive accuracy. 
However, in PolarMask~\cite{Xie_2020_CVPR, xie2021polarmask++}, contours are constructed from a set of endpoints of concentric rays emitted from the object center, thus restricting the models to handling convex shapes as well as some concave shapes with non-intersecting rays.
Another pioneering method is DeepSnake~\cite{peng2020deep} which directly regresses the coordinates of contour vertices in the Cartesian coordinate system.
Inspired by the classic Snake algorithm~\cite{kass1988snakes}, DeepSnake~\cite{peng2020deep} devises a neural network to evolve an initial contour to enclose the object boundary.
Nevertheless, the overlarge model size increases the learning difficulty and thus harms the performance.
Typically, the performance of DeepSnake~\cite{peng2020deep} drops with more than 4 times contour adjustment.
Based on this strong baseline, other methods~\cite{jetley2017straight, xu2019explicit, liang2020polytransform, Ling_2019_CVPR, Liu_2021_WACV,zhang2022e2ec} 
continue to explore a more effective contour estimation strategy.
Although they report promising performance on the challenging benchmarks~\cite{hariharan2011semantic,cordts2016cityscapes,lin2014microsoft,qi2019amodal},
their contour learning strategies are somewhat heuristic and complex.

In this work, we propose PolySnake, a new deep network architecture for generic contour-based instance segmentation.
The idea of our PolySnake is traced back to the classic Snake algorithm~\cite{kass1988snakes} that iteratively deforms an initial polygon to progressively align the object boundary by the optimization of an energy function.
Correspondingly,
our PolySnake also aims to realize effective automatic contour learning based on iterative and progressive mechanisms simultaneously,
different from existing methods.
Specifically, given an initial contour of an object instance, 
PolySnake develops a recurrent update operator to estimate the contour iteratively.
It maintains a single estimate of the contour that is progressively deformed at each iteration.
At each iteration, 
PolySnake first constructs the representation of the contour estimated at the previous iteration. 
Then, the recurrent operator takes the features as input 
and estimates the residual displacement to adjust current coarse contour to further outline the object instance.
Through the iterative refinements,
the contour progressively converges to a stable status that tightly encloses the object instance.

Our PolySnake exhibits a novel design in three aspects, discussed next.
Firstly,
PolySnake takes a neat network architecture.
Due to its recurrent design, the whole model is still lightweight and free from iteration times.
Such a lightweight architecture also ensures its real-time ability even under a relatively large number of iterations.
Secondly,
to achieve a high-quality estimation,
we present a multi-scale contour refinement module to further refine the obtained contour by aggregating fine-grained and semantic-rich feature maps.
Thirdly,
we propose a shape loss to encourage and regularize the learning of object shape, which makes the regressed contour outline the object instance more tightly.

To evaluate the effectiveness of our PolySnake, 
we conduct comprehensive experiments on several prevalent instance segmentation benchmark datasets, 
including SBD~\cite{hariharan2011semantic}, Cityscapes~\cite{cordts2016cityscapes}, COCO~\cite{lin2014microsoft}, and KINS~\cite{qi2019amodal}.
Furthermore, we validate our PolySnake in other two task scenarios, \emph{i.e.}, scene text detection (see Fig.~\ref{fig:first}(e, f)) and lane detection (see Fig.~\ref{fig:first}(g, h)).
Distinct from general instance segmentation, 
scene text detection~\cite{dai2021progressive} concentrates solely on segmenting a specific object type, that is text.
In lane detection tasks~\cite{feng2022rethinking}, the targeted lane lines inherently lack the closed-loop structure often observed in typical instance segmentation objects.
Extensive quantitative and qualitative results demonstrate the merits
of our method as well as its superiority over state-of-the-art methods.
Moreover, we
validate various design choices of PolySnake through comprehensive studies.

\begin{figure*}[t]
	\centering
	\includegraphics[width=1\linewidth]{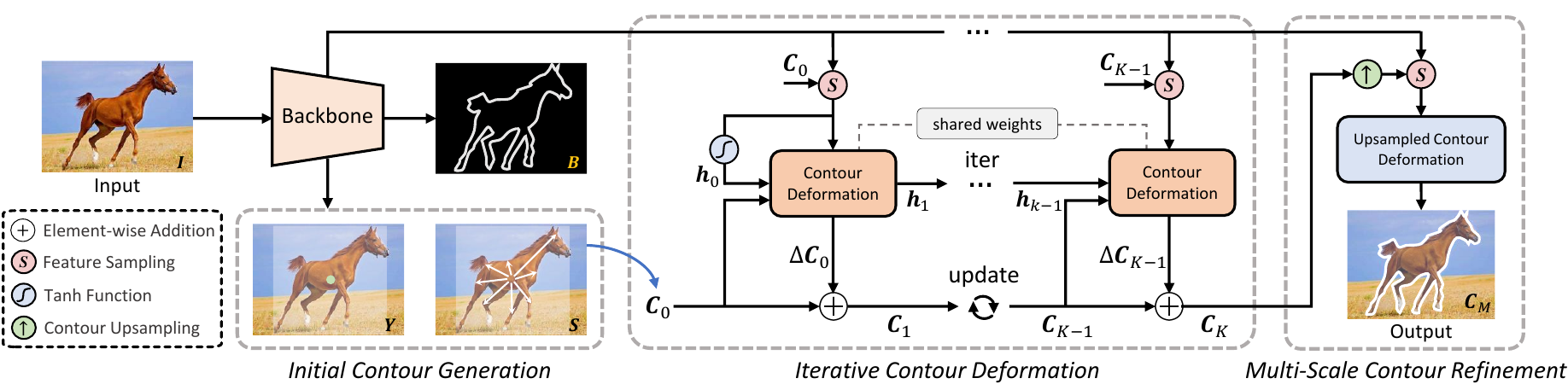}
	\vspace{-0.15in}
	\caption{
		An overview of the proposed PolySnake framework. PolySnake consists of three modules: Initial Contour Generation (ICG), Iterative Contour Deformation (ICD), and Multi-scale Contour Refinement (MCR). 
		Given an input image $\bm{I}$, ICG first initializes a coarse contour $\bm{C}_0$ based on the predicted center heatmap $\bm{Y}$ and offset map $\bm{S}$.
		Then, in ICD, 
		the obtained contour $\bm{C}_0$ is progressively deformed with $K$ iterations. After that, in MCR, we construct a large-scale but semantic-rich feature map for a further refinement
		and obtain the output contour $\bm{C}_M$.}
	\label{fig:overview}
\end{figure*}

\section{Related Work}
We mainly discuss the research on instance segmentation in two directions, including mask-based and contour-based. We discuss them in the following.

\subsection{Mask-based Instance Segmentation}
Most recent works perform instance segmentation by predicting pixel-level masks.
Some works follow the paradigm of “Detect then Segment” \cite{He_2017_ICCV, liu2018path, chen2019hybrid, Huang_2019_CVPR, shen2021dct, zhang2021refinemask,hu2021a2, tang2021look, ding2021deeply}. They usually first detect the bounding boxes and then predict foreground masks in the region of each bounding box.
Among them, the classic Mask R-CNN \cite{He_2017_ICCV} supplements a segmentation branch on Faster R-CNN \cite{ren2015faster} for per-pixel binary classification in each region proposal.
To enhance the feature representation, PANet \cite{liu2018path} proposes bottom-up path augmentation to enhance information propagation based on Mask R-CNN \cite{He_2017_ICCV}, and A$^2$-FPN \cite{hu2021a2} strengthens pyramidal feature representations through attention-guided feature aggregation.
To improve the segmentation parts, Mask Scoring R-CNN \cite{Huang_2019_CVPR} adds a MaskIoU head to learn the completeness of the predicted masks. HTC \cite{chen2019hybrid} and RefineMask \cite{zhang2021refinemask} integrate cascade into instance segmentation by interweaving detection and segmentation features \cite{chen2019hybrid} or fusing the instance features obtained from different stages \cite{zhang2021refinemask}. Recently, DCT-Mask \cite{shen2021dct} introduces DCT mask representation to reduce the complexity of mask representation, and BPR \cite{tang2021look} proposes the crop-then-refine strategy to improve the boundary quality.

There also exist some works that are free of bounding boxes. YOLACT \cite{bolya2019yolact} generates several prototype masks over the entire image and predicts a set of coefficients for each instance to combine them. BlendMask \cite{chen2020blendmask} directly predicts 2D attention map for each proposal on the top of the FCOS detector \cite{tian2019fcos}, and combines them with ROI features.
Besides, recent attempts \cite{wang2020solov2, tian2020conditional, zhang2021k, guo2021sotr} have also applied dynamic convolution kernels trained online, which are then performed on feature maps to generate instance masks.

\subsection{Contour-based Instance Segmentation}
Contour-based methods aim to predict a sequence of vertices of object boundaries, which are usually more lightweight than pixel-based methods. Some methods \cite{xu2019explicit,xie2021polarmask++,Xie_2020_CVPR} use polar-representation to predict contours directly.
Among them, 
PolarMask \cite{Xie_2020_CVPR,xie2021polarmask++} extends classic detection algorithm FCOS \cite{tian2019fcos} to instance segmentation, which adds an additional regression head to predict distances from the center point to each boundary vertex. 
However, performances of such methods are usually limited when dealing with objects with some complex concave shapes.
ESE-Seg \cite{xu2019explicit} utilizes Chebyshev polynomials~\cite{mason2002chebyshev} to approximate the shape vector and adds an extra branch on YOLOv3 \cite{redmon2018yolov3} to regress the coefficients of Chebyshev polynomials.

Some other works \cite{peng2020deep,liang2020polytransform,Liu_2021_WACV,tang2021contourrender,zhang2022e2ec} apply Cartesian coordinate representation for vertices, and regress them towards ground-truth object boundaries. 
Typically, DeepSnake \cite{peng2020deep} first initializes the contours based on the box predictions, and then applies several contour deformation steps for segmentation. Based on DeepSnake\cite{peng2020deep}, Dance \cite{Liu_2021_WACV} improves the matching scheme between predicted and target contours and introduces an attentive deformation mechanism.
ContourRender \cite{tang2021contourrender} uses DCT coordinate signature and applies a differentiable renderer to render the contour mesh.
Eigencontours~\cite{park2022eigencontours} proposes a contour descriptor based on low-rank approximation and then incorporates the eigencontours into an instance segmentation framework.
Recently, E2EC~\cite{zhang2022e2ec} applies a novel learnable contour initialization architecture and shows remarkable performance.

Distinct from existing methodologies, our PolySnake model introduces an innovative iterative and progressive mechanism for learning object contours, thereby achieving precise and robust estimations across a variety of objects.

\section{METHODOLOGY}
An overview of the proposed PolySnake is shown in Fig.~\ref{fig:overview}.
Echoing the methodology of the classic Snake algorithm, PolySnake employs an iterative approach to contour deformation, aiming to precisely align with the object boundary.
We develop a recurrent architecture that allows a single estimate of the contour progressively updated at each iteration, and further refine the obtained contour at a larger image scale.
Through iterative refinements, the contour steadily envelops the object, eventually stabilizing at a consistent state.
For the convenience of understanding, 
we divide the framework into three modules,
including (1) initial contour generation, (2) iterative contour deformation, and (3) multi-scale contour refinement.
Each module is discussed in detail in the subsequent sections.

\subsection{Initial Contour Generation}
The \underline{I}nitial \underline{C}ontour \underline{G}eneration (ICG) module is designed to construct preliminary contours for object instances.
Leveraging the principles of CenterNet~\cite{zhou2019objects},
both DeepSnake~\cite{peng2020deep} and Dance~\cite{Liu_2021_WACV} initialize a contour from the detected bounding box with a rectangle,
whereas E2EC~\cite{zhang2022e2ec}
adopts a direct regression of an ordered vertex set to form the initial contour.
In our PolySnake, we initialize the coarse contours based on the basic setting of E2EC~\cite{zhang2022e2ec}, described next.

Specifically,
given an input RGB image~$\bm{I} \in \mathbb{R}^{H \times W \times 3}$ with width $W$ and height $H$, 
we first inject it to a backbone network~\cite{yu2018deep}, and obtain a feature map $\bm{F} \in \mathbb{R}^{\frac{H}{R} \times \frac{W}{R} \times D}$, where $R=4$ is the output stride and $D$ denotes the channel number.
Then, as shown in Fig.~\ref{fig:overview}, 
the initial contour is generated with three parallel branches.
The first branch predicts a center heatmap $\bm{Y} \in[0,1] ^{\frac{H}{R} \times \frac{W}{R} \times C}$, where $C$ is the number of object categories.
The second branch regresses an offset map $\bm{S} \in \mathbb{R}^{\frac{H}{R} \times \frac{W}{R} \times 2N_v}$, where $N_v$ denotes the vertex number to form an object contour.
Let $(x_i, y_j)$ denote the center coordinate of an object,
then its contour can be obtained by adding the predicted offset $\{(\Delta x ^k, \Delta y ^k) | k=1,2,...,N_v\}$ at $(x_i, y_j)$ of $\bm{S}$.
Moreover,
we supplement an extra branch. It predicts a category-agnostic boundary map $\bm{B} \in \mathbb{R}^{\frac{H}{R} \times \frac{W}{R}}$,
in which each value is a classification confidence (boundary/non-boundary).
Our motivation is that such a branch can help to extract the more fine-grained features for the perception of object boundaries.
Note that this branch is only used in the training, which will not cause an extra computation burden for inference.
During inference, the instance center can be obtained efficiently based on the max pooling operation~\cite{zhou2019objects} on the heatmap $\bm{Y}$.
Finally, we obtain the initial contour $\bm{C}_0 \in \mathbb{R}^{N_v \times 2}$ used for the following refinements.

\subsection{Iterative Contour Deformation}\label{sec:ICD}
Given the initial contour of an object instance, the \underline{I}terative \underline{C}ontour \underline{D}eformation (ICD) module maintains a single estimation of object contour, which is refined iteratively.
In this way, the initial contour finally converges to a stable state tightly enclosing the object.
The architecture of the ICD module is neat, which contributes to the low latency of our method even with a relatively large number of iterations.

The workflow of ICD module is illustrated in Fig.~\ref{fig:overview}.
Given the feature map $\bm{F} \in \mathbb{R}^{\frac{H}{R} \times \frac{W}{R} \times D}$ extracted by the backbone network and the initial contour $\bm{C}_0 = \{\bm{p}_0^1, \bm{p}_0^2, ..., \bm{p}_0^{N_v}\} \in \mathbb{R}^{N_v \times 2}$ of an object, 
we deform the object contour iteratively and output the sequence $\{\bm{C}_1, \bm{C}_2,  ..., \bm{C}_K\}$,  
where $\bm{C}_k = \{\bm{p}_k^1, \bm{p}_k^2, ..., \bm{p}_k^{N_v}\}$ is the predicted object contour with $N_v$ vertices at the $k^{th}$ iteration, 
and $K$ is the total iteration number. 
Note that the two channels of the contour $\bm{C}_k \in \mathbb{R}^{N_v \times 2}$ denote the horizontal and the vertical coordinates of $N_v$ vertices, respectively.                                
After $K$ iterations, we produce the high-quality object contour $\bm{C}_K$. 

\begin{figure}[t]
	\centering
	\includegraphics[width=1\linewidth]{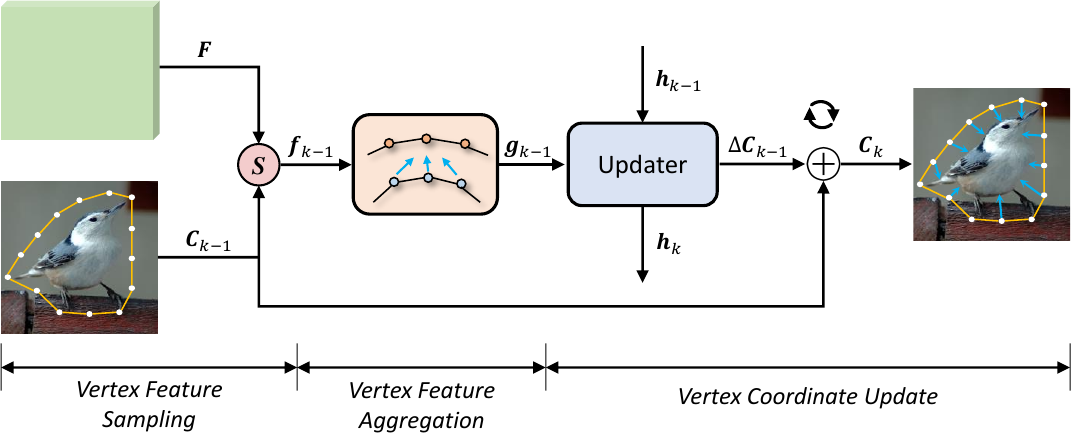}
	\caption{An illustration of the $k^{th}$ iteration in the ICD module. ``$S$" denotes vertex feature sampling based on bilinear interpolation. ``$+$" represents element-wise addition for deforming the current contour.}
	\label{fig:fea}
\end{figure}

As shown in Fig.~\ref{fig:fea}, 
we illustrate the contour update process at the $k^{th}$ iteration.
Given the feature map $\bm{F} \in \mathbb{R}^{\frac{H}{R} \times \frac{W}{R} \times D}$ and the contour $\bm{C}_{k-1}$ predicted at the ${(k-1)}^{th}$ iteration,
we first sample the vertex-wise feature and then aggregate them for deforming the current contour.  
We divide the contour update process in ICD module into three sub-processes, 
including (1) vertex-wise feature sampling, (2) vertex feature aggregation, and (3) vertex coordinate update, as described in the following.

\smallskip
\textbf{Vertex-wise feature sampling.}
Given the feature map $\bm{F} \in \mathbb{R}^{\frac{H}{R} \times \frac{W}{R} \times D}$ of image $\bm{I}$ and the contour $\bm{C}_{k-1} \in \mathbb{R}^{N_v \times 2}$ with $N_v$ vertices $\{\bm{p}_{k-1}^1, \bm{p}_{k-1}^2, ..., \bm{p}_{k-1}^{N_v}\}$ predicted at the ${(k-1)}^{th}$ iteration, 
we first retrieve features of these $N_v$ vertices. 
Specifically, given a vertex $\bm{p}_{k-1}^i \in \mathbb{R}^{2}$, we sample its feature vector $\bm{F}(\bm{p}_{k-1}^i) \in \mathbb{R}^{D}$ on the feature map $\bm{F}$ based on the bilinear interpolation. 
Note that according to classic STN~\cite{jaderberg2015spatial}, we can compute the gradients to the input $\bm{F}$ and $\bm{C}_{k-1}$ for backpropagation, thus the module can be trained in an end-to-end manner.
After that, we concatenate the feature vectors of the total $N_v$ vertices and produce the vertex feature $\bm{f}_{k-1} \in \mathbb{R}^{N_v \times D}$ at the ${(k-1)}^{th}$ iteration.

\smallskip
\textbf{Vertex feature aggregation.}
This module constructs the contour-level representation for $\bm{C}_{k-1}$ by fusing the vertex features $\bm{f}_{k-1}$.
Following~\cite{peng2020deep, Liu_2021_WACV, zhang2022e2ec, dai2021progressive}, we use the circle-convolution to fuse the feature of the contour.
The circle-convolution operation firstly joins up the head and tail of the input sequence and then applies a standard 1-D convolution on it,
which is applicable to capture features of a contour with a polygon topology. 
In our implementation, we first stack eight circle-convolution layers with residual skip connections for all the layers.
Then, we concatenate the features from all the layers and forward them through three $1\times1$ convolutional layers.
Finally, we explicitly concatenate the output features with contour coordinate $\bm{C}_{k-1} \in \mathbb{R}^{N_v \times 2}$, 
obtaining the contour representation $\bm{g}_{k-1} \in \mathbb{R}^{N_v \times D_v}$. 

\begin{figure}[t]
	\centering
	\includegraphics[width=0.98\linewidth]{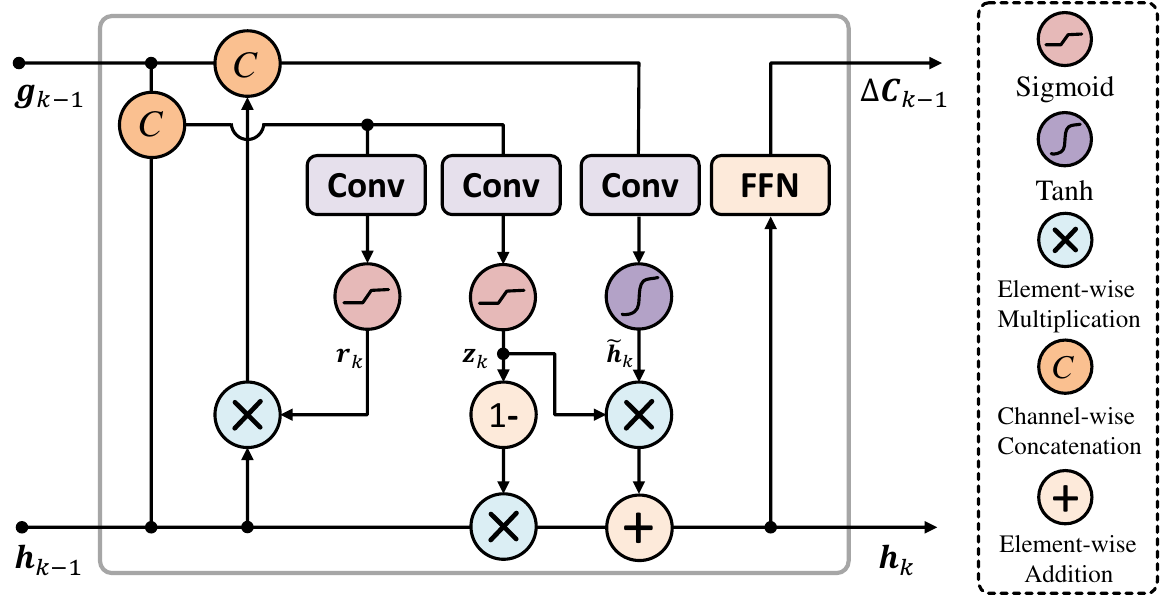}
	\caption{Inner structure of the GRU-based vertex coordinate updater.}
	\label{fig:gru}
\end{figure} 

\smallskip
\textbf{Vertex coordinate update.} 
To make the contour enclose the object tightly, we introduce a recurrent update operator that iteratively updates the currently estimated contour.
Fig.~\ref{fig:gru} illustrates the structure of our GRU-based update operator.
It is a gated activation unit based on the GRU cell~\cite{cho2014properties},
in which the fully connected layers are replaced with the 1-D convolutional layers.
At the $k^{th}$ iteration, it takes the contour representation $\bm{g}_{k-1} \in \mathbb{R}^{N_v \times D_v}$ as well as the hidden state $\bm{h}_{k-1} \in \mathbb{R}^{N_v \times D_v}$ as input, differentiates them, 
and outputs the hidden states $\bm{h}_k \in \mathbb{R}^{N_v \times D_v}$ as follows,
\begin{gather}
	\bm{z}_k=\sigma(Conv([\bm{h}_{k-1},\bm{g}_{k-1}],\bm{W}_z)), \\
	\bm{r}_k=\sigma(Conv([\bm{h}_{k-1},\bm{g}_{k-1}],\bm{W}_r)), \\
	\widetilde{\bm{h}}_k=tanh(Conv([\bm{r}_k\odot \bm{h}_{k-1},\bm{g}_{k-1}],\bm{W}_h)),  \\
	\bm{h}_k=(1-\bm{z}_k)\odot \bm{h}_{k-1}+\bm{z}_k\odot \widetilde{\bm{h}}_k,
\end{gather}
where $\odot$ denotes element-wise multiplication; $\bm{W}_z$, $\bm{W}_r$, and $\bm{W}_h$ are the weights for the corresponding gates.
Note that the initial hidden state $\bm{h}_{0}$ is initialized as the contour feature $\bm{g}_0 \in \mathbb{R}^{N_v \times D_v}$ that is processed by the $tanh$ activation function.

After that, two convolutional layers with a ReLU activation follow $\bm{h}_k \in \mathbb{R}^{N_v \times D_v}$ to produce the residual displacement $\Delta \bm{C}_{k-1} \in \mathbb{R}^{N_v \times 2}$,
which is used to update the coordinate of current contour $\bm{C}_{k-1}$ as follows,
\begin{equation}
	\bm{C}_k = \bm{C}_{k-1} + \Delta \bm{C}_{k-1}. \label{update}
\end{equation}
After $K$ iterations, we obtain the contour prediction $\bm{C}_{K} \in \mathbb{R}^{N_v \times 2}$ tightly enclosing an object instance, as shown in Fig.~\ref{fig:overview}.

\subsection{Multi-scale Contour Refinement} 
The output contour $\bm{C}_{K}$ is obtained based on the retrieved contour feature from the $\frac{1}{R}$ scale feature map $\bm{F} \in \mathbb{R}^{\frac{H}{R} \times \frac{W}{R} \times D}$.
Note that we set ${R} = 4$ by default, following DeepSnake~\cite{peng2020deep} and E2EC~\cite{zhang2022e2ec}.
However, the feature map $\bm{F}$ is semantically strong, but the high-frequency features on object boundaries are inadequate after all the downsampling and upsampling operations in the network, which brings difficulties to localize the object contours.
To this end,
we develop the MCR module, aiming to further refine the obtained $\bm{C}_K$ by aggregating a large-scale but semantic-rich feature map.

\begin{figure}[t]
	\centering
	\vspace{0.07in}
	\includegraphics[width=0.95\linewidth]{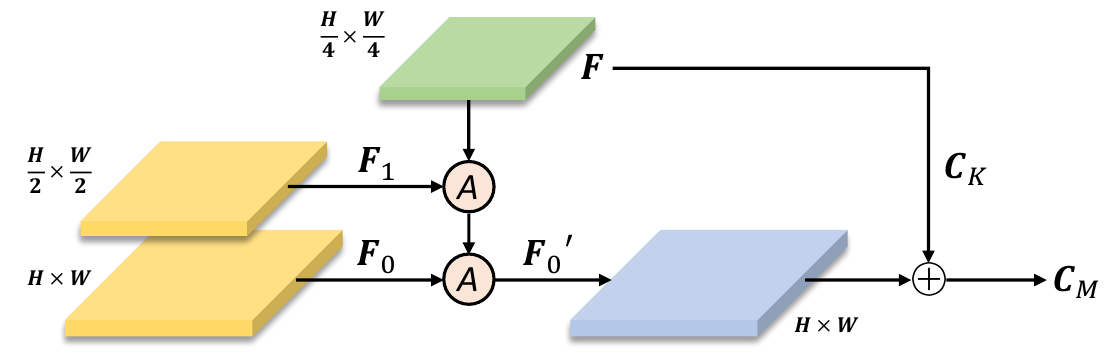}
	\caption{An overview of the multi-scale contour refinement module. ``A'' denotes the feature fusion process, following FPN~\cite{lin2017feature}.}
	\label{fig:fpn}
	\vspace{-0.07in}
\end{figure}

Specifically, we first take two additional low-level large-scale feature maps from the backbone network (\emph{i.e.}, $\bm{F}_0 \in \mathbb{R}^{H \times W \times D_0}$ and $\bm{F}_1 \in \mathbb{R}^{\frac{H}{2} \times \frac{W}{2} \times D_1}$).
Note that the high-frequency signals are reserved in $\bm{F}_0$ and $\bm{F}_1$.
Thereafter, as shown in Fig.~\ref{fig:fpn}, to aggregate them with the semantically stronger feature map $\bm{F}$, 
we build a simple feature pyramid network~\cite{lin2017feature} to construct feature map $\bm{F}_0^{'} \in \mathbb{R}^{H \times W \times D_2}$ used for following contour refinement.

Based on the feature map $\bm{F}_0^{'}$, we further deform $\bm{C}_K$ to align the object boundary.
The contour deformation process is similar to that in the ICD module, except that the GRU-based recurrent update operator is removed.
Concretely, given the feature map $\bm{F}_0^{'}$ and the current contour $\bm{C}_K$, 
we sample the per-vertex features, aggregate them, and directly predict per-vertex offset to update the contour $\bm{C}_K$ with a fully connected layer.
Finally, we obtain the refined contour estimation, termed as $\bm{C}_{M} \in \mathbb{R}^{N_v \times 2}$,
as illustrated in Fig.~\ref{fig:overview}.

\subsection{Training Objectives}\label{loss_fun_lab}
The training is divided into two stages.
Concretely,
the ICG and ICD modules are first optimized with the objective as follows,
\begin{equation}\label{equ:loss_total}
	\mathcal{L} = \mathcal{L}_{\text{ICG}} + \mathcal{L}_{\text{ICD}}.
\end{equation}
Then, we freeze the models and optimize the MCR module with objective $\mathcal{L}_{\text{MCR}}$.
In the following,
we separately elaborate them.

\smallskip
(1) $\mathcal{L}_{\text{ICG}}$ consists of three terms as follows,
\begin{equation}
	\mathcal{L}_{\text{ICG}} = \mathcal{L}_{\text{Y}} + \mathcal{L}_{\text{S}} + \mathcal{L}_{\text{B}},
\end{equation}
where $\mathcal{L}_{\text{Y}}$ and $\mathcal{L}_{\text{S}}$ are the loss on the predicted center heatmap $\bm{Y}$ and offset map $\bm{S}$.
We follow the classic detector CenterNet~\cite{zhou2019objects} to compute $\mathcal{L}_{\text{Y}}$ and $\mathcal{L}_{\text{S}}$ by only changing the output size of the offset map $\bm{S}$ from ${\frac{H}{R}\times \frac{W}{R} \times 2N_v}$ to ${\frac{H}{R} \times \frac{W}{R} \times 2}$ to directly regress the initial contour with $N_v$ vertices.
Besides, $\mathcal{L}_{\text{S}}$ is the cross-entropy loss between the predicted boundary map $\bm{B}$ and its given ground truth $\hat{\bm{B}}$ as follows,
\begin{equation}
	\mathcal{L}_{B} = -\sum_{i=1}^{N_p}\left[\hat{\bm{y}_i}\log(\bm{y}_i)+(1-\hat{\bm{y}_i})\log(1-\bm{y}_i)\right],
\end{equation}
where $N_p=\frac{H}{4}\times \frac{W}{4}$ is the number of pixels in map $\bm{B}$, $\hat{\bm{y}_i} \in [0, 1]$ and $\bm{y}_i \in \{0, 1\}$ denote the ground-truth and the predicted confidence score in $\hat{\bm{B}}$ and $\bm{B}$, respectively.

\begin{figure}[t]
	\centering
	\includegraphics[width=1\linewidth]{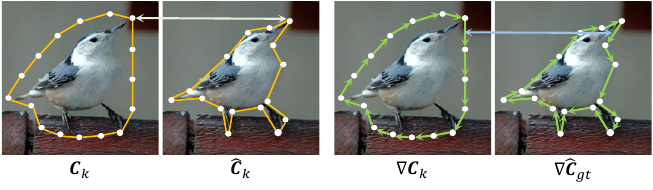}
	\vspace{-0.2in}
	\caption{An illustration of the coordinate regression loss $\mathcal{L}_{R}^{(k)}$ (left) and the proposed shape loss $\mathcal{L}_{P}^{(k)}$ (right). The loss $\mathcal{L}_{R}^{(k)}$ measure the coordinate distance between the predicted contour $\bm{C}_k$ and GT contour $\hat{\bm{C}}_{gt}$, while $\mathcal{L}_{R}^{(k)}$ encourages the learning of object shape.}
	\label{fig:shape}
\end{figure}

\smallskip
(2) $\mathcal{L}_{\text{ICD}}$ is calculated by accumulating the loss over all $K$ iterations as follows,
\begin{equation}\label{equ:loss_icd}
	\mathcal{L}_{\text{ICD}}=\sum_{k=1}^K \lambda ^{K-k} (\mathcal{L}_{R}^{(k)} + \alpha  	\mathcal{L}_{P}^{(k)}), 
\end{equation}
where $\mathcal{L}_{R}^{(k)}$ and $\mathcal{L}_{P}^{(k)}$ are the coordinate regression loss and our proposed contour shape loss at the $k^{th}$ iteration, $\lambda$ is the temporal weighting factor, and $\alpha$ is the weight of contour shape loss. Here $\lambda$ is less than 1, which means the weight of the loss increases exponentially with the iteration.
In Fig.~\ref{fig:shape}, we illustrate the coordinate regression loss $\mathcal{L}_{R}^{(k)}$ (left) and the proposed shape loss $\mathcal{L}_{P}^{(k)}$ (right).
Specifically,
we compute the coordinate regression loss by calculating the smooth $L_1$ distance~\cite{ren2015faster} between the predicted contour $\bm{C}_k =\{\bm{p}_k^1, \bm{p}_k^2, ..., \bm{p}_{k}^{N_v}\}$ and the ground truth $\hat{\bm{C}_{gt}} =\{\hat{\bm{p}}^1, \hat{\bm{p}}^2, ..., \hat{\bm{p}}^{N_v}\}$ as follows,
\begin{equation}
	\mathcal{L}_{R}^{(k)}=\sum_{n=1}^{N_v} \text{smooth}_{L1}(\bm{p}_{k}^n, \hat{\bm{p}}^n).
\end{equation}
Furthermore, we denote the offsets of adjacent points in each contour $\nabla \bm{C}_k = \{ \Delta \bm{p}_{k}^{2\rightarrow1}, \Delta \bm{p}_{k}^{3\rightarrow2}, ..., \Delta \bm{p}_{k}^{{N_v}\rightarrow {N_v}-1},\Delta \bm{p}_{k}^{1\rightarrow {N_v}}\}$ as the shape representation of contour $\bm{C}_k$. Note that since the contour is a closed polygon, the offset between the first and the last points is included in $\nabla \bm{C}_k$. Then, the proposed shape loss $\mathcal{L}_{P}^{(k)}$ is defined as the smooth $L_1$ distance~\cite{ren2015faster} between $\nabla \bm{C}_k$ and the ground truth shape representation $\nabla \hat{\bm{C}}_{gt}$, computed as follows,
\begin{equation}
	\mathcal{L}_{P}^{(k)}=\sum_{n=1}^{N_v} \text{smooth}_{L1}(\Delta \bm{p}_{k}^{{n+1}\rightarrow{n}}, \Delta \hat{ \bm{p}}^{{n+1}\rightarrow{n}}).
\end{equation}
The shape loss encourages the learning of object shape and makes the regressed contour enclose the object tightly.

\smallskip
(3) $\mathcal{L}_{\text{MCR}}$ is calculated as the smooth $L_1$ distance~\cite{ren2015faster} between the refined contour $\bm{C}_M=\{\bm{p}_M^1, \bm{p}_M^2, ..., \bm{p}_M^{N_v}\}$ and its given ground truth $\hat{\bm{C}_{gt}}$ as follows,
\begin{equation}
	\mathcal{L}_{\text{MCR}}=\sum_{n=1}^{N_v} \text{smooth}_{L1}(\bm{p}_M^n, \hat{\bm{p}}^n).
\end{equation}

\section{Experiments}

\subsection{Datasets and Metrics}
Extensive experiments have been conducted on three widely recognized datasets for instance segmentation, as well as two additional datasets specifically focused on scene text detection and lane detection, described in the following.

\smallskip             
\textbf{SBD}~\cite{hariharan2011semantic} dataset consists of 5,623 training and 5,732 testing images, spanning 20 distinct semantic categories. 
It utilizes images from the PASCAL VOC \cite{everingham2010pascal} dataset, but re-annotates them with instance-level boundaries. 
In our study of the SBD dataset, we report the performance of prior works and our PolySnake based on the metrics 2010 VOC AP$_{vol}$ \cite{hariharan2014simultaneous}, AP$_{50}$, and AP$_{70}$. 
AP$_{vol}$ is computed as the average of average precision (AP) with nine thresholds from 0.1 to 0.9.

\smallskip             
\textbf{Cityscapes}~\cite{cordts2016cityscapes} dataset is intended for assessing the performance of vision algorithms of semantic urban scene understanding. The dataset contains $5000$ images with high-quality annotations of 8 semantic categories. 
They are further split into $2975$, $500$, and $1525$ images for training, validation, and testing, respectively.
Results are evaluated in terms of the AP metric averaged over all categories of the dataset.

\smallskip                
\textbf{KINS}~\cite{qi2019amodal} dataset is specifically constructed for amodal instance segmentation~\cite{li2016amodal}, 
which adopts images from the KITTI~\cite{geiger2013vision} dataset and annotates them with instance-level semantic annotation.
The dataset contains 7474 training and 7517 testing images.
We evaluate our approach on seven object categories with the AP metric.

\setlength{\tabcolsep}{2.0mm}        
\begin{table}[t]
	\small
	\centering
	\caption{Ablations of the initial contour generation module on the SBD val set~\cite{hariharan2011semantic}.
		$\bm{C}_0$ and $\bm{C}_K$ are the output contour of the ICG and ICD modules, respectively. Settings used in our final model are underlined.}
        \vspace{-0.01in}
	\begin{tabular}{cccccc}
          \toprule
		Experiment & Contour & Method & AP$_{vol}$ & AP$_{50}$ & AP$_{70}$ \\
            \midrule
		\multirow{4}*{Boundary Map} & \multirow{2}*{$\bm{C}_0$} &  \underline{w/} & \textbf{49.5} & \textbf{61.2} & 31.9  \\
		& & w/o & 49.3 & 61.0 & \textbf{32.2}  \\
		\cmidrule(lr){2-6}
		& \multirow{2}*{$\bm{C}_K$} &  \underline{w/} & \textbf{59.7} & \textbf{66.7} & \textbf{54.8}  \\
		& & w/o & 59.3 & 66.6 & 54.4 \\
        \midrule
		
		\multirow{3}*{Vertex Number} & \multirow{3}*{$\bm{C}_K$}  & $N=64$   & 58.9 & 66.0 & 54.1   \\
		& & \underline{$N=128$} & \textbf{59.7} & \textbf{66.7} & \textbf{54.8}  \\
		& & $N=192$            & 59.2 & 66.2 & 54.3 \\
		
        \bottomrule
	\end{tabular}
	\label{tab:aba1}
\end{table}        

\setlength{\tabcolsep}{0.91mm}        
\begin{table}[t]
	\small
	\centering
	\caption{Ablations of the iterative contour deformation module on the SBD val set~\cite{hariharan2011semantic}. 
		Settings used in our final model are underlined.}
          \vspace{-0.01in}
	\begin{tabular}{lcccccc}

        \toprule
		Experiment & Method & AP$_{vol}$ & AP$_{50}$ & AP$_{70}$ & Para. (M) & FPS \\
	
        \midrule
	\multirow{2}*{Contour Feature}  & $\bm{f}_{k}$  & 58.6 & 65.7 & 53.2 & 20.3 & 44.2 \\
		
		& \underline{$\bm{g}_{k}$} & \textbf{59.7} & \textbf{66.7} & \textbf{54.8} & 22.0 & 28.8 \\
        \midrule
		
		\multirow{2}*{Update Unit}  & \underline{ConvGRU}  & \textbf{59.7} & \textbf{66.7} & \textbf{54.8} & 22.0 & 28.8  \\
		& ConvLSTM  & 59.5 & 66.4 & 54.6 & 22.1 & 24.4 \\
        \midrule
		
		\multirow{2}*{Shared Weights}  & \underline{w/} & \textbf{59.7} & \textbf{66.7} & 54.8 & 22.0 & -  \\
		& w/o  & 59.6 & 66.5 & \textbf{55.1} & 31.0 & - \\
        \midrule

		\multirow{2}*{Supervised iters}  & \underline{[1,$K$]} & \textbf{59.7} & \textbf{66.7} & \textbf{54.8} & 22.0 & -  \\
		& $\{K\}$  & 58.8 & 66.2 & 54.0 & 22.0 & - \\

        \midrule
  
		\multirow{2}*{Shape Loss}  & \underline{w/} & \textbf{59.7} & \textbf{66.7} & \textbf{54.8} & - & - \\
		& w/o  & 59.4 & 66.2 & \textbf{54.8} & - & - \\
   \bottomrule
		
	\end{tabular}
	
	\label{tab:aba2}
\end{table}    

\smallskip               
\textbf{COCO}~\cite{lin2014microsoft} dataset stands as a substantial and challenging resource for instance segmentation. It encompasses an extensive collection of images, comprising 115,000 for training, 5,000 for validation, and 20,000 for testing, spanning 80 diverse object categories. A notable feature of the COCO dataset is its inclusion of everyday objects and human figures within its imagery. In our research, we employ the COCO Average Precision (AP) metric as the standard for evaluating the efficacy of other methods and our method.

\smallskip
\textbf{CTW1500}~\cite{liu2019curved} dataset, designed for scene text detection, specifically addresses the challenge of detecting curved texts. It contains 1,500 images, split into 1,000 training and 500 testing images, and features a total of 10,751 text instances, including a significant proportion of curved text. The dataset's variety includes images sourced from the internet, image libraries, and phone cameras.

\smallskip   
\textbf{CULane}~\cite{pan2018spatial} is a large challenging lane detection dataset. It consists of 9 types of scenarios, including normal, crowd, curve, dazzling light, night, no line, shadow, and arrow.
Additionally, it includes a diverse range of environments, capturing both urban and highway settings. In the CULane~\cite{pan2018spatial} dataset, the maximum number $M$ of lane lines in the scene is 4.

\setlength{\tabcolsep}{4.5mm}        
\begin{table}[t]
	\small
	\centering
	\caption{Performance of the selected iterations during inference on the SBD val set~\cite{hariharan2011semantic}. $\bm{C}_0$ is output initial contour of the ICG module. The running efficiency is evaluated on a single RTX 2080Ti GPU. The contour used as the final output of the ICD module is underlined.}
	\begin{tabular}{ccccc}
          \toprule
		Contour & AP$_{vol}$ & AP$_{50}$ & AP$_{70}$ & FPS  \\
		\midrule
		$\bm{C}_0$ & 49.5 & 61.2 & 31.9 & 51.5 \\
		$\bm{C}_1$ & 56.1 & 65.6 & 49.2 & 48.4 \\
		$\bm{C}_2$ & 58.5 & 66.3 & 53.8 & 45.1 \\
		$\bm{C}_3$ & 59.3 & 66.5 & 54.4 & 37.9 \\
		$\bm{C}_4$ & 59.6 & 66.6 & 54.7 & 34.6 \\
		$\bm{C}_5$ & 59.7 & 66.6 & 54.7 & 30.5 \\
		\underline{$\bm{C}_6$} & 59.7 & 66.7  & 54.8 & 28.8 \\
		$\bm{C}_7$ & 59.7 & 66.7  & 54.9 & 26.1 \\
		$\bm{C}_8$ & 59.7 & 66.7  & 54.9 & 23.6 \\
		\midrule
		$\bm{C}_M$ & \textbf{60.0} & \textbf{66.8}  & \textbf{55.3} & 24.6 \\
		\bottomrule
	\end{tabular}
	\label{tab:aba3}
\end{table}       

\begin{figure}[t]
	\centering
	\includegraphics[width=0.94\linewidth]{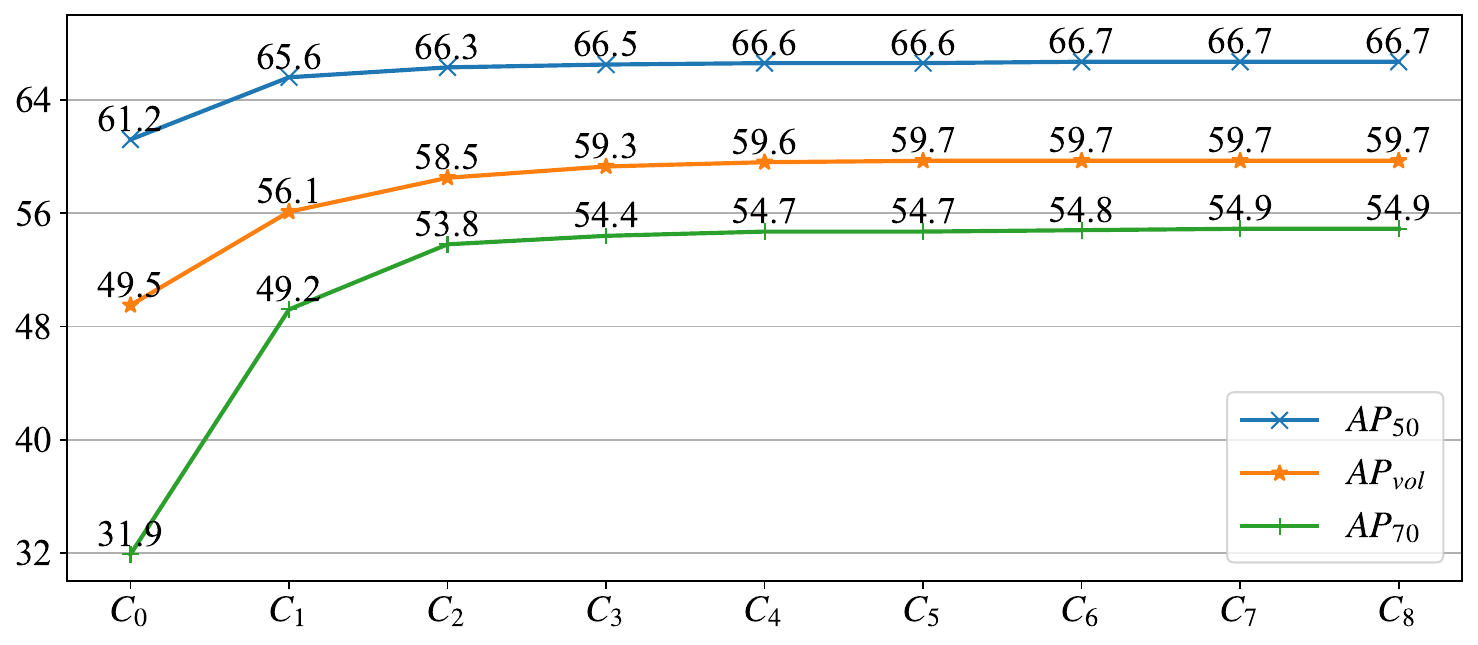}
	\vspace{-0.05in}
	\caption{Performance plot of the selected iterations during inference on the SBD val set~\cite{hariharan2011semantic}. $\bm{C}_0$ is output initial contour of the ICG module.}
	\label{fig:aplot}
\end{figure}

\subsection{Implementation Details}
By default, we set the vertex number $N_v=128$ to form an object contour.
The downsampling ratio $R$ and the channel dimension $D$ for feature map $\bm{F}$ are set as 4 and 64, respectively. 
The channel dimension $D_v$ of contour feature $\bm{g}_{k-1}$ is 66.
The channel dimension of feature map $\bm{F}_0$, $\bm{F}_1$, $\bm{F}_0^{'}$ in the MCR module are 32, 16, and 8, respectively. 
We set $\lambda=0.8$ and $\alpha=1$ in Equation~\eqref{equ:loss_icd} to calculate loss $\mathcal{L}_{\text{ICD}}$, respectively.
During training, we set the iteration number $K$ as 6 for the SBD~\cite{hariharan2011semantic}, Cityscapes~\cite{cordts2016cityscapes}, COCO~\cite{lin2014microsoft}, and KINS~\cite{qi2019amodal} dataset. 
By default, the iteration number during the inference is the same as that for the training.

\subsection{Ablation Study}
In this section, we conduct ablations to verify the effectiveness of the main components in our proposed PolySnake,
including the initial contour generation, the iterative contour deformation, and the multi-scale contour refinement.
Following previous methods~\cite{peng2020deep,zhang2022e2ec}, all the ablations are studied on the SBD~\cite{hariharan2011semantic} dataset with 20 semantic categories, which well evaluates the algorithm capability to handle various object shapes.
Note that in the ablations, our baseline network does not include the multi-scale contour refinement module.

\begin{figure}[t]
	\centering
	\includegraphics[width=0.96\linewidth]{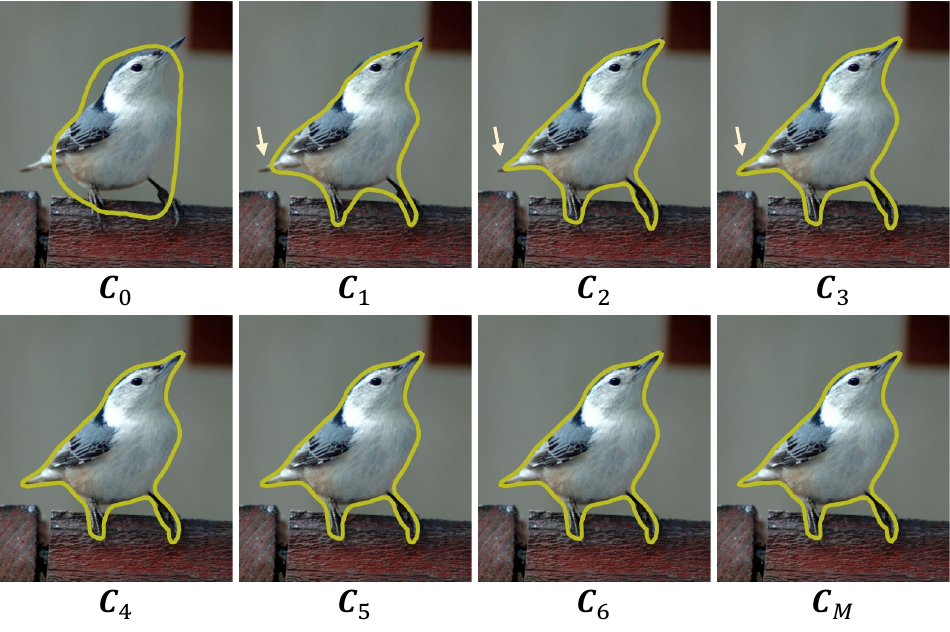}
	\vspace{-0.1in}
	\caption{Example results of the whole contour deformation process on the SBD val set~\cite{hariharan2011semantic}.
		PolySnake progressively refines the contour.
		The contour finally converges to a stable position, which tightly encloses the object instance. Zoom in for a better view.}
	\label{fig:iter_vis_sbd}
\end{figure} 

\begin{figure}[t]
	\centering
	\includegraphics[width=0.96\linewidth]{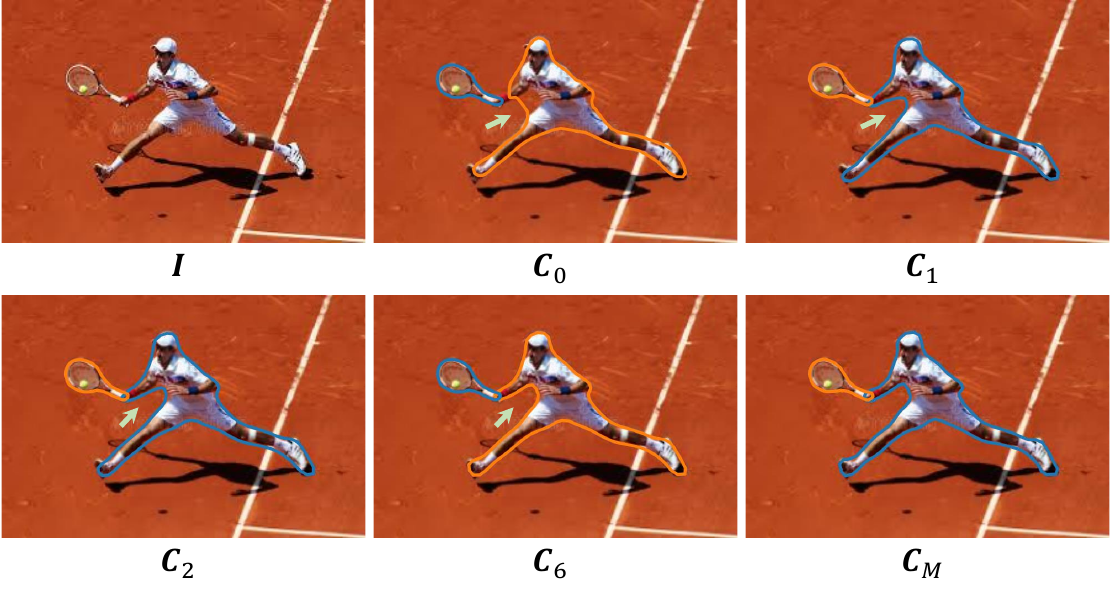}
	\vspace{-0.1in}
	\caption{Example results of the contour deformation process on the COCO~\cite{lin2014microsoft} dataset.
		PolySnake progressively refines the contour to tightly enclose the object instance. Zoom in for a better view.}
	\label{fig:iter_vis_coco}
\end{figure}

\smallskip
\textbf{Initial contour generation.}
We first validate the learning of category-agnostic boundary map $\bm{B}$ for the contour initialization.
As shown in Table~\ref{tab:aba1},
with the learning of boundary map $\bm{B}$,
the performance of contour $\bm{C}_0$ and $\bm{C}_K$ are both slightly improved. 
This is because the learning of category-agnostic boundary map $\bm{B}$ helps to extract a more fine-grained feature map $\bm{F}$, which improves the perception of object contour.

\begin{figure}[b]
	\centering
	\subfigure[Performance]{
		\label{flow_loss}
		\includegraphics[width=0.47\linewidth]{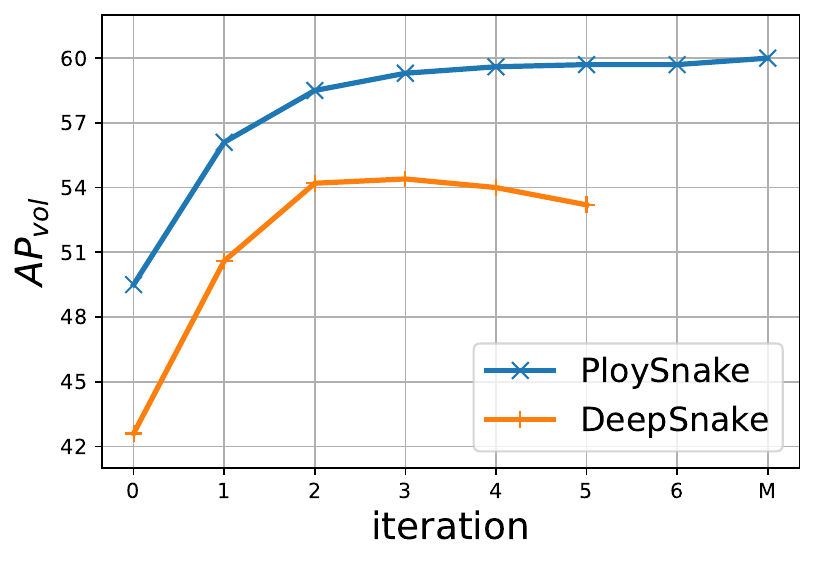}}
	\subfigure[Parameters]{
		\label{mask_loss}
		\includegraphics[width=0.47\linewidth]{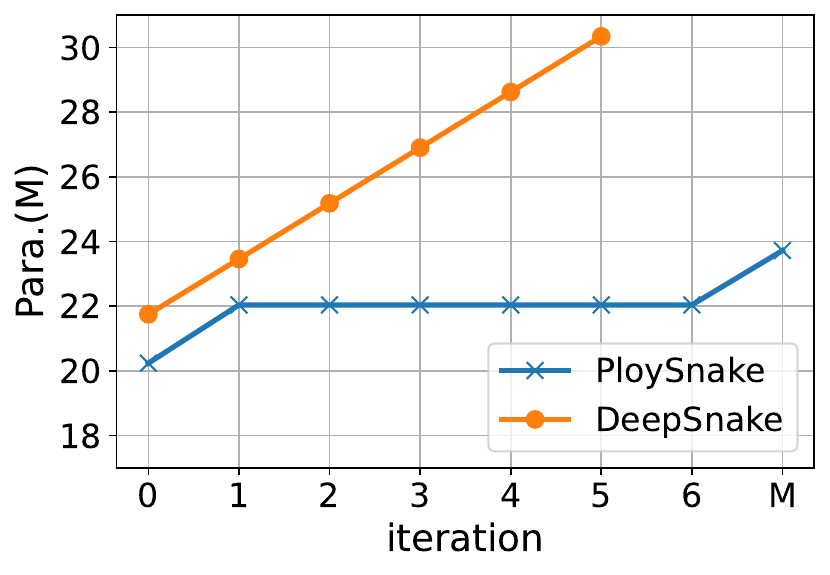}}
	\caption{Results of DeepSnake~\cite{peng2020deep} and our PolySnake on the performance (left) and the parameter numbers (right) at different time steps on the SBD val set~\cite{hariharan2011semantic}. ``$0$'' denotes the initial contour for DeepSnake~\cite{peng2020deep} and PolySnake. ``$M$'' represents the MCR module in PolySnake.}
	\label{fig:per_para}
\end{figure}  

Furthermore, we test the impact of the vertex number $N_v$ to construct a contour.
The default vertex number $N_v$ of PolySnake is 128.
We study the other settings of vertex number $N_v$ as 64 and 192, respectively.
The results are reported in Table~\ref{tab:aba1}. It reveals that 128 vertices are enough to represent the contour of an object instance. In contrast, sampling more vertices (192) leads to worse performance. 
A likely reason is that it is difficult for the contour feature extraction module to obtain a strong representation from an overlong sequence.

\smallskip
\textbf{Iterative contour deformation.}
We first validate the contour feature extraction used to construct the contour-level representation $\bm{g}_{k}$ of the estimated contour $\bm{C}_{k}$. 
Specifically, as shown in Table~\ref{tab:aba2}, 
we train a baseline model that directly feeds the sampled vertex feature $\bm{f}_{k-1}$ to the recurrent updater (see Fig.~\ref{fig:fea}).
In other words, the baseline model does not aggregate the feature of the sampled vertices.
With the aggregated vertex feature $\bm{g}_{k-1}$,
the result shows a performance gain of 1.1 AP$_{vol}$.
This improvement could be ascribed to the fact that the aggregated feature $\bm{g}_{k-1}$ encodes stronger context information of the current contour $\bm{C}_{k-1}$,
which is differentiated and processed by the recurrent updater to estimate further refinement.

In the ICD module, the default update unit is the ConvGRU, to which an alternative is the ConvLSTM, a modified version of the standard LSTM~\cite{hochreiter1997long}.
As shown in Table~\ref{tab:aba2}, while the ConvLSTM shows comparable performance, the ConvGRU produces slightly higher efficiency and accuracy.

By default, our PolySnake shares the weights across the total $K$ iterations. An alternative version is to learn each vertex feature aggregation module and vertex coordinate update module (see in Sec.~\ref{sec:ICD}) with a distinctive set of weights. 
In Table~\ref{tab:aba2}, with the unshared weights, the performances are slightly worse while the parameter number significantly increases.
For one thing, this can be attributed to the increased training difficulty of the large model size, which is verified in DeepSnake~\cite{peng2020deep}.
For another, our effective information transmission mechanism across iterations and the supervision at each iteration maintain the performance (see Table~\ref{tab:aba2}), different from DeepSnake~\cite{peng2020deep}.

We further compare our contour deformation pipeline with the classic DeepSnake~\cite{peng2020deep}.
In Fig.~\ref{fig:per_para}, we present the results of PolySnake and DeepSnake~\cite{peng2020deep} in terms of the performance (left) and the parameter numbers (right) at different iterations on the SBD val set~\cite{hariharan2011semantic}. 
In DeepSnake~\cite{peng2020deep}, the iteration number is fixed as three by default. 
On the one hand, simply stacking more refinement modules ($>3$) introduces extra training and does not necessarily improve the performance;
on the other hand, the parameter number increases linearly with more iterations.
In contrast, for our PolySnake, the superior performance is obtained after convergence with only a lightweight model, indicating the stronger and stabler contour-deforming ability of our method.

During training, we introduce a contour shape loss in Sec.~\ref{loss_fun_lab}. 
To validate its effectiveness, we evaluate the performance of our approach with and without it. The results, presented in Table~\ref{tab:aba2}, reveal that incorporating the contour shape loss yields an improvement of 0.3 in AP$_{vol}$.
We attribute this improvement to the fact that the proposed contour shape loss helps the learning of the object shape, which makes the estimated contour enclose the object tightly.

\setlength{\tabcolsep}{3mm} 
\begin{table}[t]
	\centering
	\caption{Comparison of the AP$_{vol}$, AP$_{50}$, and AP$_{70}$ performances on the \textbf{SBD} val set~\cite{hariharan2011semantic}. ``$^\dag$'' means that the multi-scale contour refinement module is used.}
	\begin{tabular}{lcccc}
          \toprule
		Method & Venue & AP$_{vol}$ & AP$_{50}$ & AP$_{70}$  \\
        \midrule
		MNC \cite{dai2016instance} & \emph{CVPR'16} & - & 63.5 & 41.5  \\
		FCIS \cite{li2017fully} & \emph{CVPR'17} & - & 65.7 & 52.1  \\
		STS \cite{jetley2017straight} & \emph{CVPR'17} & 29.0 & 30.0 & 6.5  \\
		ESE-20 \cite{xu2019explicit} & \emph{ICCV'19} & 35.3 & 40.7 & 12.1  \\
		DeepSnake \cite{peng2020deep} & \emph{CVPR'20} & 54.4 & 62.1 & 48.3  \\
		DANCE \cite{Liu_2021_WACV} & \emph{WACV'21} & 56.2 & 63.6 & 50.4 \\
		EigenContours \cite{park2022eigencontours} & \emph{CVPR'22} & - & 56.5 & -  \\
		E2EC \cite{zhang2022e2ec} & \emph{CVPR'22} & 59.2 & 65.8 & 54.1  \\
		\midrule
		PolySnake  & - & 59.7 & 66.7 & 54.8 \\
		PolySnake$^\dag$   & - & \textbf{60.0} & \textbf{66.8} & \textbf{55.3} \\
		\bottomrule
	\end{tabular}
	\label{table:sbd}
\end{table}

\setlength{\tabcolsep}{2.5mm}   
\begin{table}[t]
	\centering
	\caption{Performance comparison with other instance segmentation methods on the \textbf{COCO} \texttt{val} and \texttt{test}-\texttt{dev} set~\cite{lin2014microsoft}. ``$^\dag$'' means that the multi-scale contour refinement module is added.}
	\begin{tabular}{lcccc}
          \toprule
		\multirow{2}{*}{Method} & \multirow{2}{*}{Venue} & \multirow{2}{*}{Backbone} & AP  & AP  \\
		&    &  & (val) & (test-dev)  \\
        \midrule
		ESE-Seg \cite{xu2019explicit} & \emph{ICCV'19} & Dark-53  & 21.6 & - \\
		YOLACT~\cite{bolya2019yolact} & \emph{ICCV'19} & R-50-FPN & - & 28.2 \\
		YOLACT++~\cite{bolya2020yolact++} & \emph{TPAMI'20} & R-50-FPN & - & 34.1 \\
		CenterMask~\cite{lee2020centermask} & \emph{CVPR'20} & R-50-FPN & - & 32.9 \\
		PolarMask \cite{Xie_2020_CVPR} & \emph{CVPR'20} & R-50-FPN & - & 32.1      \\
		PolarMask++ \cite{xie2021polarmask++} & \emph{TPAMI'21} & R-101-FPN & - & 33.8      \\
		DeepSnake \cite{peng2020deep} & \emph{CVPR'20} & DLA-34 & 30.5 & 30.3 \\
		E2EC \cite{zhang2022e2ec} & \emph{CVPR'22} & DLA-34 & 33.6 & 33.8  \\
		\midrule
		PolySnake  & - & DLA-34 & 34.4 & 34.5  \\
		PolySnake$^\dag$   & - & DLA-34 & \textbf{34.8} & \textbf{34.9} \\
    \bottomrule
	\end{tabular}
	\label{tab:coco}
\end{table}

\setlength{\tabcolsep}{6.2mm} 
\begin{table}[t]
	\centering
	\caption{Performance comparison on the \textbf{KINS} test set~\cite{qi2019amodal}. ``*'' denotes with ASN proposed in~\cite{Qi_2019_CVPR}. ``$^\dag$'' means that the multi-scale contour refinement module is added.}
	\begin{tabular}{lcc}
          \toprule
		Method & Venue  & AP \\
            \midrule
		MNC \cite{dai2016instance} & \emph{CVPR'16} & 18.5   \\
		FCIS \cite{li2017fully} & \emph{CVPR'17}  & 23.5   \\
		ORCAN \cite{follmann2019learning} & \emph{WACV'19} & 29.0 \\
		Mask R-CNN \cite{He_2017_ICCV} & \emph{ICCV'17}  & 30.0   \\
		Mask R-CNN * \cite{Qi_2019_CVPR} & \emph{CVPR'19}  & 31.1   \\
		PANet \cite{liu2018path} & \emph{CVPR'18}  & 30.4   \\
		PANet * \cite{Qi_2019_CVPR} & \emph{CVPR'19}  & 32.2   \\
		VRS$\&$SP \cite{xiao2021amodal} & \emph{AAAI'21} & 32.1 \\
		ARCNN \cite{zeng2022amodal} & \emph{AI'22} & 32.9 \\
		DeepSnake \cite{peng2020deep} & \emph{CVPR'20}  & 31.3   \\
		E2EC \cite{zhang2022e2ec} & \emph{CVPR'22} & 34.0 \\
		\midrule
		PolySnake  & -  & 35.0 \\
		PolySnake$^\dag$   & -  & \textbf{35.2} \\
            \bottomrule
	\end{tabular}
	\label{table:kins}
\end{table}

\setlength{\tabcolsep}{5mm} 
\begin{table}[t]
	\centering
	\caption{Performance comparison on the \textbf{CTW1500} test set~\cite{liu2019curved}. $\mathcal{R}$, $\mathcal{P}$, and $\mathcal{F}$ respectively denote the recall rate, precision, and F1-score~\cite{liao2020real}.}
	\begin{tabular}{lccc}
          \toprule
		Method  & $\mathcal{R}$ & $\mathcal{P}$ & $\mathcal{F}$ \\
            \midrule
            LOMO~\cite{zhang2019look}  & 76.5 & 85.7 & 80.8 \\
            PSENet-1s~\cite{wang2019shape} & 79.7 & 84.8 & 82.2 \\
            CRAFT~\cite{baek2019character} & 81.1 & 86.0 & 83.5 \\
            DBNet~\cite{liao2020real}   &  80.2 & 86.9 & 83.4 \\
            PAN~\cite{wang2019efficient} & 81.2 & 86.4 & 83.7 \\
            TextDragon~\cite{feng2019textdragon}  &  82.8 & 84.5 & 83.6 \\
            DRRG~\cite{zhang2020deep} &  \textbf{83.0} & 85.9 & 84.5 \\
            CounterNet~\cite{wang2020contournet}  & 84.1 & 83.7 & 83.9 \\
            PCR~\cite{dai2021progressive}  & 82.3 & 87.2 &  \textbf{84.7} \\
            TextBPN~\cite{zhang2021adaptive}  & 80.6 & 87.7 & 84.0 \\
            PAN++~\cite{wang2021pan++} & 81.1 & 87.1 & 84.0 \\ 
		\midrule
		PolySnake   & 81.5 & \textbf{88.1} & \textbf{84.7}   \\
            \bottomrule
	\end{tabular}
	\label{table:ctw}
\end{table}

\begin{figure*}[t]
	\centering
	\includegraphics[width=0.98\linewidth]{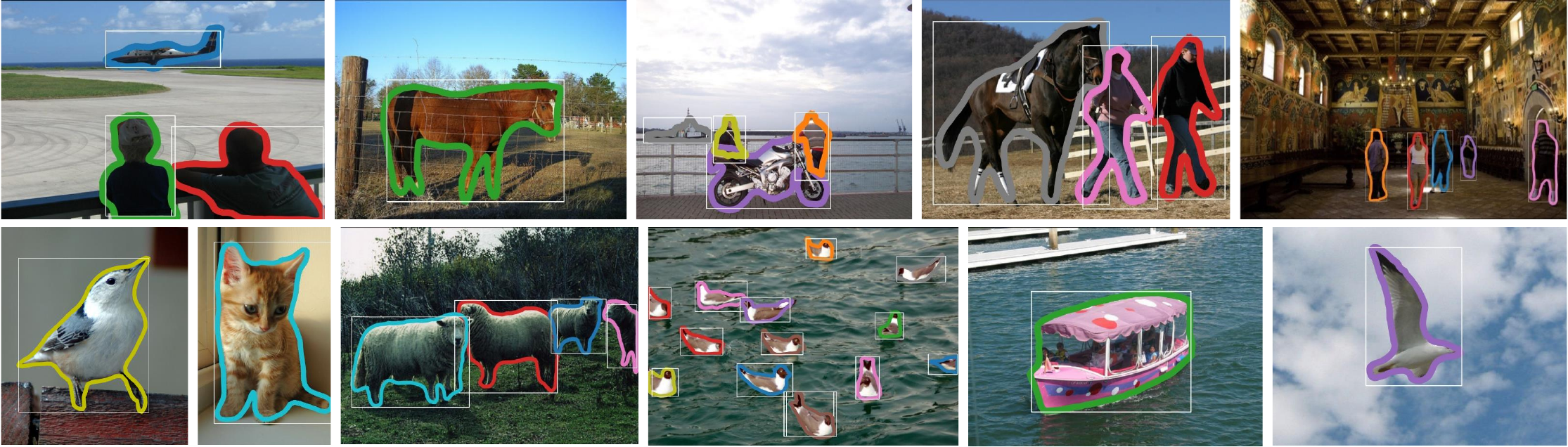}
 	\vspace{-0.05in}
	\caption{Qualitative results of PolySnake on the SBD val set~\cite{hariharan2011semantic}. Our method can segment object instances correctly in most cases, even when they are in some complex backgrounds or the detected boxes can not cover the objects tightly. It is best viewed in color.}
	\label{fig:sbd_vis}
\end{figure*}

\smallskip
\textbf{Iterative and progressive mechanism.}
To provide a more specific view of the iterative contour deformation process, 
we present the performance at the selected iteration in the inference stage.
As shown in Table~\ref{tab:aba3}, the main contour deformation occurs in the top $3$ iterations, 
while the later iterations further progressively fine-tune the result.
We also present the performance plots in Fig.~\ref{fig:aplot}. As we can see, the results are getting better with iterations and the superior performance is obtained after convergence.
These results verify the effectiveness of our method.
In our final model, we set the iteration number $K=6$ to strike a balance between the accuracy and the running efficiency.
Besides, the contour $\bm{C}_M$ achieves an improvement of 0.5 AP$_{70}$ compared with the contour $\bm{C}_6$,
which validates the MCR module that further refines the contour at the full scale.

As shown in Fig.~\ref{fig:iter_vis_sbd} and Fig.~\ref{fig:iter_vis_coco}, 
we further visualize the contour deformation process with two samples.
It can be seen that, during the contour deformation process, 
the predicted contours are progressively refined and finally converge to relatively steady positions enclosing the object tightly.

\begin{figure*}[t]
	\centering
	\includegraphics[width=0.98\linewidth]{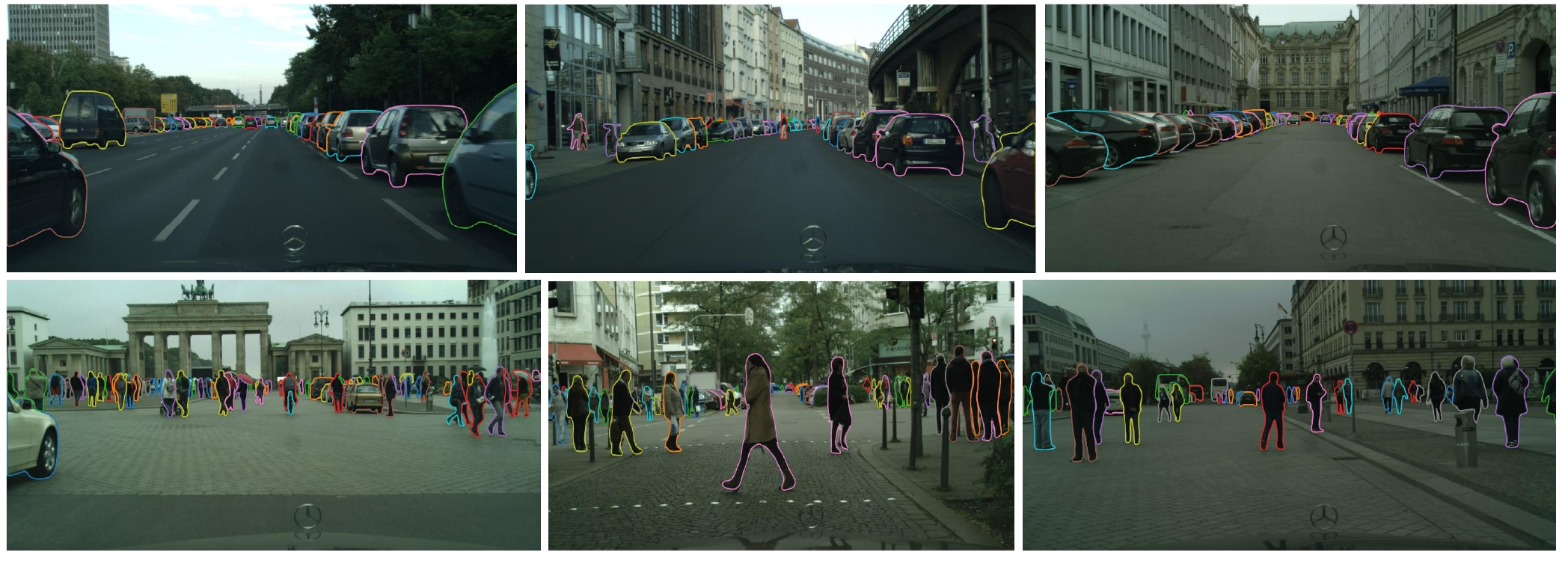}
 	\vspace{-0.1in}
	\caption{Qualitative results of PolySnake on the Cityscapes test set~\cite{cordts2016cityscapes}. Our method is able to correctly segment objects in most cases, even when they are in complex street scenes. It is best viewed in color.}
	\label{fig:city_vis}
\end{figure*}

\begin{figure*}[!ht]
	\centering
	\includegraphics[width=0.98\linewidth]{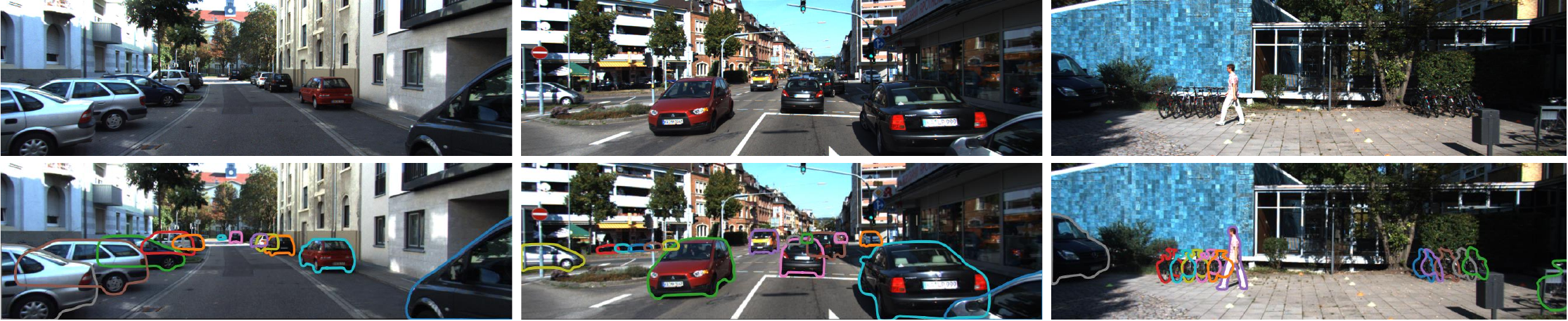}
	\caption{Qualitative results of PolySnake on the KINS test set~\cite{qi2019amodal}. The top and bottom rows are the input images and their segmentation results, respectively. PolySnake can correctly segment objects with occluded parts in complex scenes. It is best viewed in color.}
	\label{fig:kins_vis}
  	\vspace{-0.05in}
\end{figure*}

\setlength{\tabcolsep}{1.94mm}  
\begin{table*}[t]
    \centering
    \caption{Comparison results on the \textbf{Cityscapes} val (``AP [\texttt{val}]'' column) and test (remaining columns) sets~\cite{cordts2016cityscapes}. Using only fine training data, PolySnake achieves superior performance on both the val and test sets. The inference time of method~\cite{peng2020deep,Liu_2021_WACV,zhang2022e2ec} and PolySnake are measured on a single RTX 2080Ti GPU, while we report the results of the other methods from their original papers. The best and second-best results in total and each category are highlighted in \textbf{bold} and \underline{underlined}, respectively.}
	\begin{tabular}{l|c|c|c|cccccccccc} 
          \toprule
		Method & Training data & FPS & AP [\texttt{val}] & AP & AP$_{50}$ 
		& Person & Rider & Car & Truck & Bus & Train & Mcycle & Bicycle \\
	\midrule
		SGN~\cite{liu2017sgn} & \texttt{fine} + \texttt{coarse} & 0.6 & 29.2 & 25.0 & 44.9 & 21.8 &	20.1 &	39.4 &	24.8 &	33.2 & 30.8 &	17.7 &	12.4 \\
		Mask R-CNN~\cite{He_2017_ICCV} & \texttt{fine} & 2.2 & 31.5 & 26.2 & 49.9 & 30.5 & 23.7 & 46.9 & 22.8 & 32.2 & 18.6 & 19.1 & 16.0 \\
		GMIS~\cite{liu2018affinity} & \texttt{fine} + \texttt{coarse} & - & - & 27.6 & 49.6 & 29.3 & 24.1 & 42.7 & 25.4 & 37.2 & \textbf{32.9} & 17.6 & 11.9 \\
		Spatial~\cite{neven2019instance} & \texttt{fine} & 11 & - & 27.6 & 50.9 & 34.5 & 26.1 & 52.4 & 21.7 & 31.2 & 16.4 & 20.1 & 18.9 \\
		PANet~\cite{liu2018path} & \texttt{fine} & $<$1 & 36.5 & 31.8 & 57.1 & 36.8 & 30.4 & 54.8 & 27.0 & 36.3 & 25.5 & \underline{22.6} & \textbf{20.8} \\
		UPSNet~\cite{xiong2019upsnet} & \texttt{fine} + \texttt{COCO} & 4.4 & 37.8 & 33.0 &59.6 & 35.9 & 27.4 & 51.8 & 31.7  & \underline{43.0} & 31.3 & \textbf{23.7} & 19.0 \\
		SSAP ~\cite{gao2019ssap} & \texttt{fine} & - & 37.3 & 32.7 & 51.8 & 35.4 & 25.5 & 55.9 & \textbf{33.2} & \textbf{43.9} & \underline{31.9} & 19.5 & 16.2 \\
		PolygonRNN++~\cite{acuna2018efficient} & \texttt{fine} & - & - & 25.5 & 45.5 & 29.4 & 21.8 & 48.3 & 21.1 & 32.3 & 23.7 & 13.6 & 13.6 \\
		DeepSnake~\cite{peng2020deep} & \texttt{fine} & 4.6 & 37.4 & 31.7 & 58.4 & 37.2 & 27.0 & 56.0 & 29.5 & 40.5 & 28.2 & 19.0 & 16.4 \\
		DANCE~\cite{Liu_2021_WACV} & \texttt{fine} & 6.3 & 36.7 & 31.2 & 57.7 & 38.1 & 27.3 & 54.0 & 27.5 & 37.4 & 27.7 & 21.6 & 16.2 \\
		E2EC~\cite{zhang2022e2ec} & \texttt{fine} & 4.9 & 39.0 & 32.9 & 59.2 & 39.0 & 27.8 & 56.0 & 28.5 & 41.2 & 29.1 & 21.3 & \underline{19.6} \\
		\midrule
		PolySnake & \texttt{fine} & 4.8 & \underline{39.8} & \underline{34.3} & \underline{61.0} & \underline{41.3} & \underline{31.8} & \underline{58.4} & 31.9 & 42.4 & 28.6 & 22.4 & 17.9 \\
		PolySnake$^\dag$ & \texttt{fine} & 4.2 & \textbf{40.2} & \textbf{34.6} & \textbf{61.2} & \textbf{42.2} & \textbf{32.8} & \textbf{59.2} & \underline{32.0} & 42.5 & 27.4 & \underline{22.6} & 18.4\\
        \bottomrule
	\end{tabular}
	\label{table:cityscapes}
	\vspace{0.02in}
\end{table*}

\subsection{Comparison with State-of-the-art Methods}
\textbf{Performance on SBD.}
PolySnake is implemented in Pytorch~\cite{paszke2017automatic}.
During training, we train the ICG and ICD modules end-to-end for 200 epochs.
Then, we freeze their weights and train the MCR module for 50 epochs.
The batch size is 40 for both training stages and we use Adam~\cite{kingma2014adam} as the optimizer.
We use 4 and 2 RTX 2080Ti GPUs for the two training stages, respectively.
The learning rate starts from $1e{-4}$ and then decays based on the StepLR strategy. The network is trained and tested at a single scale of $512 \times 512$, following~\cite{peng2020deep,zhang2022e2ec}, following DeepSnake~\cite{peng2020deep}.

In Table~\ref{table:sbd}, we compare the proposed PolySnake with other contour-based methods~\cite{jetley2017straight, xu2019explicit, peng2020deep, Liu_2021_WACV,park2022eigencontours, zhang2022e2ec} on the SBD val set~\cite{hariharan2011semantic}.
Our approach achieves 60.0 AP$_{vol}$ when the
multi-scale contour refinement strategy is adopted.
Compared with the classic DeepSnake~\cite{peng2020deep}, PolySnake yields 5.6 AP$_{vol}$, 4.7 AP$_{50}$, and 7.0 AP$_{70}$ improvements.
Besides,
PolySnake outperforms the recent state-of-the-art method E2EC~\cite{zhang2022e2ec} by 0.8 AP$_{vol}$, 1.0 AP$_{50}$, and 1.2 AP$_{70}$.
We present some qualitative segmentation results of our method in Fig.~\ref{fig:sbd_vis}.

\smallskip
\textbf{Performance on COCO.}
For the challenging COCO~\cite{lin2014microsoft} dataset,
we train the ICG and ICD modules following the strategy of E2EC~\cite{zhang2022e2ec}.
Then, we freeze them and train the MCR module for another 50 epochs with the same strategy as the SBD~\cite{hariharan2011semantic} dataset.
The network is trained at a single image scale of $512 \times 512$ on 2 RTX 3090Ti GPUs.
For the evaluation, we select a model that performs best on the validation set.

As shown in Table~\ref{tab:coco}, 
PolySnake achieves 34.8 and 34.9 AP on the COCO val and test-dev set, respectively.
It outperforms the classic DeepSnake~\cite{peng2020deep} by 4.3 and 4.6 AP on the val and test-dev set, respectively.
Besides, compared with the recent state-of-the-art method E2EC~\cite{zhang2022e2ec}, it achieved a 1.2 and 1.1 AP improvement on the val and test-dev set, respectively.

\begin{figure*}[t]
    \centering
    \includegraphics[width=2\columnwidth]{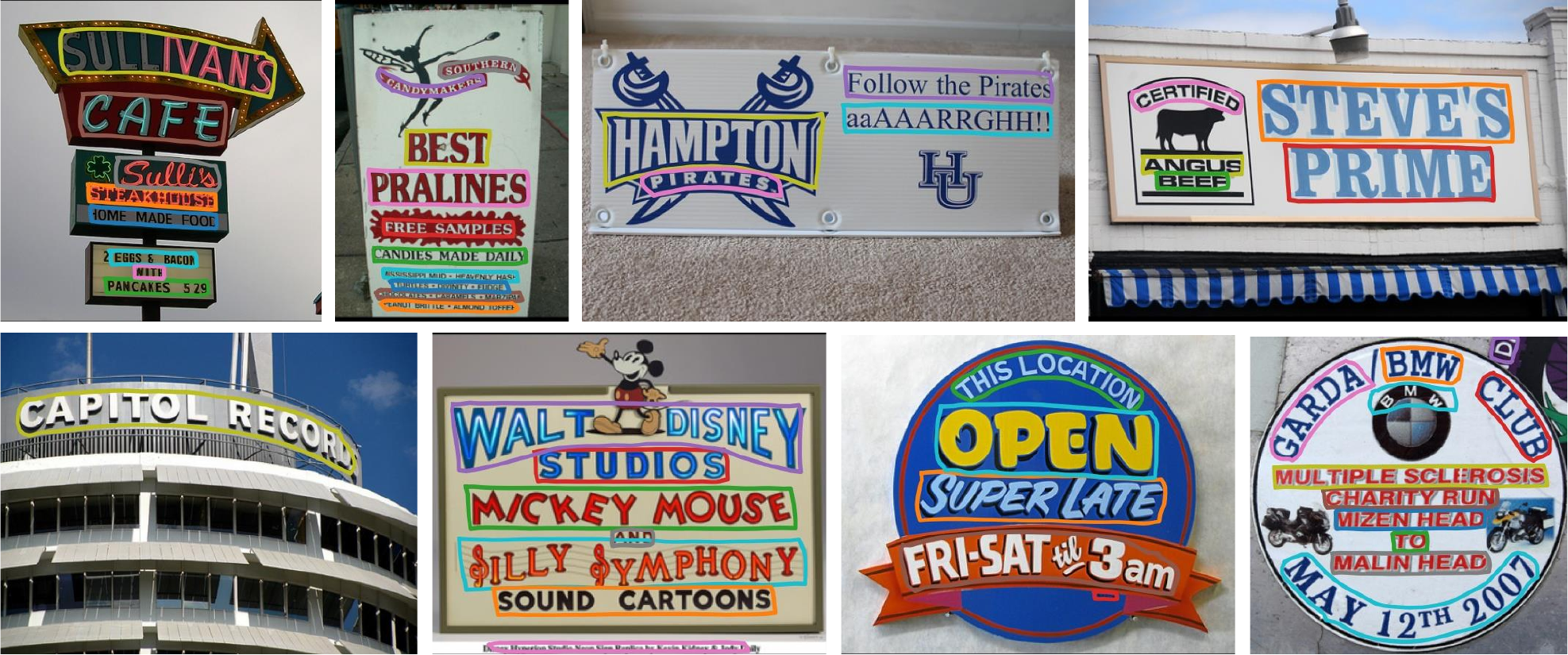}
    \caption{Visualization of scene text detection results by the PolySnake on the CTW1500 test set~\cite{liu2019curved}. Best viewed on screen.
    }
    \vspace{-0.1in}
    \label{fig:std_vis}
\end{figure*}

\smallskip
\textbf{Performance on Cityscapes.}
As suggested in DeepSnake~\cite{peng2020deep}, we apply the component detection strategy~\cite{He_2017_ICCV} to handle the fragmented instances, which are frequently occurred in the dataset.
We train the ICG and ICD modules end-to-end for 200 epochs with 16 images per batch on 4 RTX 2080Ti GPUs.
Then, their weights are frozen and we train the MCR
module for 50 epochs.
The Adam optimizer~\cite{kingma2014adam} is used,
and the learning rate schedule is the same as that for the SBD~\cite{hariharan2011semantic} dataset.
We choose the model performing best on the validation set and the evaluation is conducted at a single resolution of $1216 \times 2432$,
following DeepSnake~\cite{peng2020deep}.

As shown in Table~\ref{table:cityscapes}, 
we compare our PolySnake with other state-of-the-art methods on the Cityscapes validation and test sets~\cite{cordts2016cityscapes}. 
Using only the fine annotations, 
our approach achieves superior performances on both the validation and test sets.
We outperform DeepSnake~\cite{peng2020deep} by 3.5 AP and 3.4 AP on the validation set and test set, respectively.
Compared with recent state-of-the-art E2EC~\cite{zhang2022e2ec},
PolySnake achieves a 1.2 AP and 1.7 AP improvement on the validation and test set, respectively.
We present some segmentation results in Fig.~\ref{fig:city_vis}.

\smallskip
\textbf{Performance on KINS.}
The KINS~\cite{qi2019amodal} dataset is used for amodal instance segmentation~\cite{li2016amodal}, and is annotated with inference completion information for the occluded parts of the instances.
We first train the ICG and ICD modules for 150 epochs with the Adam optimizer~\cite{kingma2014adam}. The learning rate starts from $1e-4$ and is decayed by a factor of 0.5 at 80 and 120 epochs. 
Then, we freeze their weights and train the MCR module for another 50 epochs as the setting for the SBD~\cite{hariharan2011semantic} dataset.
Following the classic DeepSnake~\cite{peng2020deep}, the models are trained and tested at a single resolution of $512 \times 512$ and $768\times 2496$, respectively.

As shown in Table~\ref{table:kins},
the MCR module improves the performance from 35.0 AP to 35.2 AP.
We note that the improvement is somewhat marginal.
A likely reason is that the sampled vertex feature at the occluded parts can not provide the clues for further accurate refinement, different from the other three datasets~\cite{hariharan2011semantic,cordts2016cityscapes,lin2014microsoft}. 
Besides,
our PolySnake outperforms DeepSnake~\cite{peng2020deep} and E2EC~\cite{zhang2022e2ec} by 3.9 and 1.2 AP, respectively.
We present some segmentation results in Fig.~\ref{fig:kins_vis}.
It can be seen that the proposed PolySnake can correctly segment objects with occluded parts to each other in complex urban scenes.

\begin{table*}[!t]
    \caption{
    Comparison results on \textbf{CULane}~\cite{pan2018spatial} test set.
    The ``Cross'' category presents false positive results because it has no lane line in the scenario. The best and second-best results in total and each category are highlighted in \textbf{bold} and \underline{underlined}, respectively.}
    \setlength{\tabcolsep}{2.8mm} 
    \centering
    \begin{tabular}{l|c|ccccccccc} 
        \toprule
        Method & Total & Normal &  Crowded & Dazzle & Shadow & No line & Arrow & Curve & Cross & Night \\
        \midrule
        SCNN (VGG16)~\cite{pan2018spatial} & 71.60 & 90.60 & 69.70 &  58.50 & 66.90 & 43.40 &  84.10 & 64.40 & 1990 &  66.10 \\
        CurveLane-NAS-S~\cite{xu2020curvelane} & 71.40 & 88.30 & 68.60 & 63.20 & 68.00 & 47.90 & 82.50 & 66.00 & 2817 & 66.20 \\
        CurveLane-NAS-M~\cite{xu2020curvelane} & 73.50 & 90.20 & 70.50 & 65.90 & 69.30 & 48.80 & 85.70 & 67.50 & 2359 & 68.20 \\
        CurveLane-NAS-L~\cite{xu2020curvelane} &  74.80 & 90.70 & 72.30 & 67.70 & 70.10 & 49.40 & 85.80 & 68.40 & 1746 & 68.90 \\ 
        PINet (Hourglass)~\cite{ko2021key} & 74.40 & 90.30 & 72.30 &  66.30 &  68.40 & \underline{49.80} & 83.70 & 65.20 &  1427 & 67.70\\
        RESA (ResNet-34)~\cite{zheng2021resa} & 74.50 & 91.90 & 72.40 & 66.50 & 72.00 & 46.30 & 88.10 & 68.60 & 1896 & 69.80\\
        RESA (ResNet-50)~\cite{zheng2021resa} &  75.30 & 92.10 & 73.10 & 69.20 & 72.80 & 47.70 & 88.30 & \textbf{70.30} & 1503 & 69.90 \\
        LaneATT (ResNet-18)~\cite{tabelini2021keep} & 75.13 & 91.17 & 72.71 & 65.82 & 68.03 & 49.13 & 87.82 & 63.75 & 1020 & 68.58\\
        LaneATT (ResNet-34)~\cite{tabelini2021keep} & \underline{76.68} & 92.14 & \textbf{75.03} & 66.47 & \textbf{78.15} & 49.39 & \underline{88.38} & 67.72 & 1330 & 70.72\\
        LaneAF (ERFNet)~\cite{abualsaud2021laneaf} & 75.63 &  91.10 & 73.32 &  \underline{69.71} & 75.81 & \textbf{50.62} & 86.86 &  65.02 & 1844 & \underline{70.90}\\
        UFLD (ResNet-18)~\cite{qin2020ultra} & 68.40 & 87.70 &  66.00 & 58.40 & 62.80 & 40.20 & 81.00 & 57.90 & 1743 &  62.10\\
        UFLD (ResNet-34)~\cite{qin2020ultra} & 72.30 & 90.70 & 70.20 &  59.50 & 69.30 & 44.40 & 85.70 & 69.50 &  2037 &  66.70\\
        UFLDv2 (ResNet-18)~\cite{qin2022ultra} & 75.00 & 91.80 & 73.30 & 65.30 & 75.10 & 47.60 & 87.90 & 68.50 & 2075 & 70.70\\
        UFLDv2 (ResNet-34)~\cite{qin2022ultra} & 76.00 & \textbf{92.50} & 74.80 & 65.50 & 75.50 & 49.20 & \textbf{88.80} & \underline{70.10} & 1910 & 70.80\\
        BézierLaneNet (ResNet-18)~\cite{feng2022rethinking} & 73.67 & 90.22 & 71.55 & 62.49 & 70.91 & 45.30 & 84.09 &  58.98 &  996 & 68.70\\
        BézierLaneNet (ResNet-34)~\cite{feng2022rethinking} & 75.57 & 91.59 & 73.20 & 69.20 & \underline{76.74} & 48.05 & 87.16 & 62.45 & \textbf{888} & 69.90\\
        \midrule
        PolySnake (ResNet-18) & 75.99 & 91.89 & 74.34 & \textbf{70.04} & 73.44 & 48.07 & 87.34 & 68.26 & 1084 & 70.37\\
        PolySnake (ResNet-34) & \textbf{76.82} & \underline{92.49} & \underline{75.02} & 68.73 & 75.27 & 48.33 & 87.35 & 67.55 & \underline{899} & \textbf{72.30}\\
        \bottomrule
    \end{tabular}
    \vspace{0.05in}
    \label{tab:result1}
\end{table*}

\begin{figure*}[t]
    \centering
    \includegraphics[width=2\columnwidth]{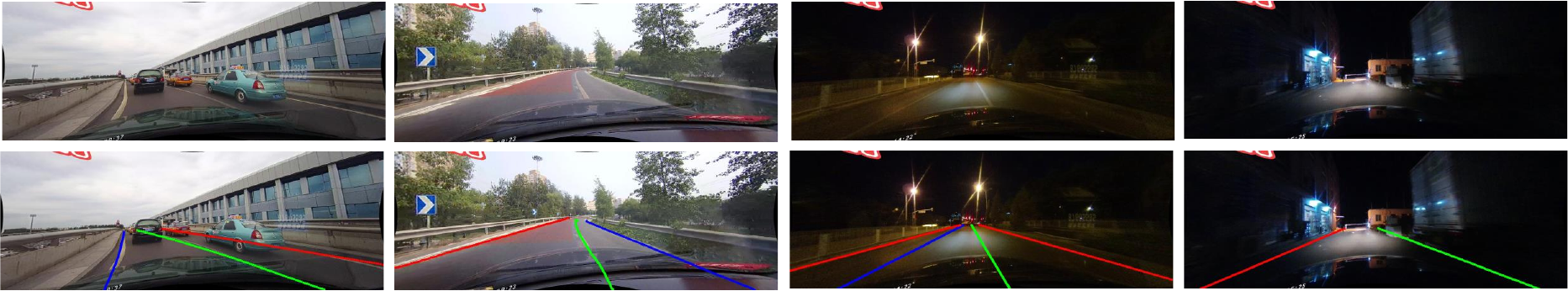}
    \caption{Visualization of lane detection results by the PolySnake on the CULane dataset~\cite{pan2018spatial}. The images in the vertical direction are the original input image and predictions, respectively. Lane markings are annotated with different colors to indicate different lanes. It is best viewed in color.
    }
    \vspace{-0.1in}
    \label{fig:vis1}
\end{figure*}

\smallskip
\textbf{Performance on CTW1500}.
For scene text detection, 
the quantitative results on the CTW1500~\cite{liu2019curved} are presented in Table~\ref{table:ctw}.
For this task, we exclusively trained the ICG and ICD modules in an end-to-end manner.
The training process spanned 210 epochs, utilizing two 2080Ti GPUs with a batch size of 24.
Note that here the vertex number $N_v$ of a contour is set as 96,
which yields the best optimal performance.

As shown in Table~\ref{table:ctw},
our PolySnake demonstrates outstanding performance. Noteworthy is its remarkable precision, a testament to the accuracy of PolySnake in contour regression. Furthermore, as illustrated in Fig.~\ref{fig:std_vis}, we further visualize the results of this test benchmark. It is observable that texts of various shapes within the images are accurately segmented. The above exemplary performance in text scenes underscores the robust generalizability of our PolySnake.

\smallskip
\textbf{Performance on CULane}.
For lane detection,
the quantitative results on the CULane~\cite{pan2018spatial} benchmark dataset are presented in Table~\ref{tab:result1}. We compare our methodology with other prominent approaches, employing ResNet-18~\cite{he2016deep} and ResNet-34~\cite{he2016deep} as backbones.
The initial lane line is constructed based on BézierLaneNet~\cite{feng2022rethinking}.
The vertex number $N_v$ of a contour is set as 100,
which yields the best optimal performance.
It is noteworthy that, in contrast to task instance segmentation and scene text detection, the representation of lane lines is characterized by a sequence of non-closed points. Hence, within the ICD module in Fig.~\ref{fig:fea}, we employ the standard 1-D conventional convolution instead of circle-convolution~\cite{peng2020deep} to perform the vertex feature aggregation.

PolySnake outperforms many segmentation-based methods on CULane~\cite{pan2018spatial}.
It also surpasses anchor-based methods, such as LaneATT~\cite{tabelini2021keep}, UFLD~\cite{qin2020ultra}, and UFLDv2~\cite{qin2022ultra}. The anchor-based methods do not perform well in the dazzle category, where our approach exhibits significantly higher accuracy.
Compared with LaneATT~\cite{tabelini2021keep} that achieves a high total score, our approach surpasses 6.41\% in the dazzle category on ResNet-18~\cite{he2016deep}.
Our approach attains the highest F1 score among curve-based approaches, with improvement of 3.15\% on ResNet-18~\cite{he2016deep} and 1.65\% on ResNet-34~\cite{he2016deep} over BézierLaneNet~\cite{feng2022rethinking}.
Besides, our approach achieves outstanding results in difficult categories, such as night.
The visualization results on CULane~\cite{pan2018spatial} are presented in Fig.~\ref{fig:vis1}.

\section{Conclusion}
In this work, we introduce PolySnake, a novel deep network architecture for contour-based instance segmentation.
PolySnake uniquely combines iterative and progressive learning mechanisms to facilitate the learning of contour estimation.
By developing a recurrent architecture, PolySnake maintains a single estimation of a contour for each object instance and progressively updates it toward the object boundary.
Through the iterative refinement, 
the contour finally
progressively converges to a stable status that tightly encloses the
object instance.
Extensive experiments are conducted on several prevalent benchmark datasets across multiple tasks.
The results reveal the effectiveness and generalizability of our PolySnake over existing advanced solutions.
For future research, we will explore improving the state-of-the-art instance segmentation methods through effective integration with PolySnake.
Moreover, it is significant to excavate our idea on other polygon or curve estimation problems in computer vision.

{
\bibliographystyle{IEEEtran}
\bibliography{IEEEabrv,egbib}
}

\vfill

\end{document}